\documentclass[11pt]{article}

\usepackage{acl}

\usepackage{times}
\usepackage{latexsym}
\usepackage{amsmath}
\usepackage{amssymb}
\usepackage{mathtools}
\usepackage{amsthm}
\usepackage{subcaption}
\usepackage{wrapfig,lipsum,booktabs}
\usepackage{hyperref}
\usepackage{url}
\usepackage{algorithm}
\usepackage{algpseudocode}
\usepackage{sidecap}
\usepackage{amsmath,bm}
\usepackage{multirow}
\usepackage{graphicx}
\usepackage{comment}
\usepackage{amssymb}
\usepackage{enumitem}
\usepackage{commath}
\usepackage{mathtools, nccmath}
\usepackage{enumitem}
\usepackage{color, colortbl}
\usepackage{tikz}
\usepackage[most]{tcolorbox}

\colorlet{colexam}{blue!90!black}
\tcbset{
  base/.style={
    empty,
    frame engine=path,
    colframe=yellow!10,
    sharp corners,
    title={Case study},
    attach boxed title to top left={yshift*=-\tcboxedtitleheight},
    boxed title style={size=minimal, top=4pt, left=4pt},
    coltitle=colexam,fonttitle=\small\bfseries\sffamily,
  }
}
\newtcolorbox[use counter=example]{myexamplea}{%
  base,
  boxed title style={overlay={
    \draw[colexam,line width=2pt,] (frame.north west)--(frame.north east);
  }},
  colback=colexam,
  overlay unbroken={
    \draw[colexam] ([yshift=-1.5pt]title.north east)--([xshift=-0.5pt, yshift=-1.5pt]title.north-|frame.east);
  },
}

\usepackage[T1]{fontenc}

\usepackage[utf8]{inputenc}

\usepackage{microtype}

\usepackage{inconsolata}

\title{\texttt{Mirror}: A Multiple-perspective 
Self-Reflection Method \\ for Knowledge-rich Reasoning}

\author{Hanqi Yan$^{1}$\thanks{\quad Equal Contribution.}\quad Qinglin Zhu$^{1*}$\quad Xinyu Wang$^{1,2}$\quad Lin Gui$^{1}$\quad  Yulan He$^{1,2,3}$
\\$^{1}$King's College London \quad $^{2}$University of Warwick \quad $^{3}$The Alan Turing Institute\\
$\{\texttt{hanqi.yan, qinglin.1.zhu, lin.1.gui, yulan.he}\}$\texttt{@kcl.ac.uk} \\
\texttt{xinyu.wang.11@warwick.ac.uk}
}

\newcommand{\outline}[1]{\textcolor{black}{#1}}

\begin{document}
\maketitle

\begin{abstract}
\outline{While Large language models (LLMs) have the capability to iteratively reflect on their own outputs, recent studies have observed their struggles with knowledge-rich problems without access to external resources. In addition to the inefficiency of LLMs in self-assessment, we also observe that LLMs struggle to revisit their predictions despite receiving explicit negative feedback. Therefore, We propose \texttt{Mirror}, a \texttt{\textbf{M}}ult\texttt{\textbf{i}}ple-pe\texttt{\textbf{r}}spective self-\texttt{\textbf{r}}eflection method for kn\textbf{\texttt{o}}wledge-rich \textbf{\texttt{r}}easoning, to avoid getting stuck at a particular reflection iteration. \texttt{Mirror} enables LLMs to reflect from multiple-perspective clues, achieved through a heuristic interaction between a Navigator and a Reasoner. It guides agents toward diverse yet plausibly reliable reasoning trajectory without access to ground truth by encouraging (1) diversity of directions generated by Navigator and (2) agreement among strategically induced perturbations in responses generated by the Reasoner. The experiments on five reasoning datasets demonstrate that \texttt{Mirror}'s superiority over several contemporary self-reflection approaches. Additionally, the ablation study studies clearly indicate that our strategies alleviate the aforementioned challenges. The code is released at \href{https://github.com/hanqi-qi/Mirror.git}{https://github.com/hanqi-qi/Mirror.git.}}
\end{abstract}
\section{Introduction}
Large Language Models (LLMs) have become an important and
flexible building block in a variety of tasks.
They can be further improved by iterative correction in many tasks~\cite{DBLP:journals/corr/abs-2303-17651,DBLP:journals/corr/abs-2305-11738,Shinn2023ReflexionLA,Pan2023AutomaticallyCL}, such as code generation, arithmetic problem solving and reasoning. During iterative refinement, the critic module, which assesses the current response and generates valuable feedback, is crucial to drive performance improvement. 

Some research shows that LLMs have self-assessment abilities~\cite{DBLP:journals/corr/abs-2303-08896,DBLP:journals/corr/abs-2303-17651}.  For example, LLMs can reject its own prediction and generate a response \textit{`I don't know'} when they are not confident about their predictions~\cite{DBLP:journals/corr/abs-2207-05221}. Empirical observations demonstrate LLMs' competence in various reasoning tasks, leading to the utilization of advanced LLMs to evaluate the predictions made by other models~\citep{DBLP:journals/corr/abs-2305-14992,DBLP:journals/corr/abs-2310-04406,liu2023training}. 
However, recent studies suggest that relying directly on LLMs' judgements is not trustworthy and can lead to failures in knowledge-rich iterative reasoning~\cite{DBLP:journals/corr/abs-2310-01798}. To guide LLMs through a reasoning loop, existing solutions either incorporate external resources to verify LLMs' outputs~\cite{DBLP:journals/corr/abs-2302-12813,DBLP:conf/iclr/YaoZYDSN023}, or train a critic module on labelled assessment datasets~\cite{DBLP:journals/corr/abs-2305-11738,zelikman2022star}. Furthermore, self-consistency is considered a robust unsupervised method to identify confident and reliable LLM outputs.

In self-refinement, the quality of generated feedback also plays a pivotal role. The Self-Refine method~\citep{DBLP:journals/corr/abs-2303-17651} introduced task-specific metrics for multifaceted feedback generation, requiring LLMs to evaluate their outputs across various aspects, such as \textit{fluency, engagement}, and \textit{relevance} for the dialogue generation task. This process often  heavily relies on human expertise, and generating effective feedback for reasoning tasks can be even more difficult as it is obscure to define the essential attributes for different problems. Providing overly general feedback fails to guide LLMs toward generating better outputs in subsequent iterations.

\begin{figure*}
    \includegraphics[width=0.92\textwidth,trim={0 10 12 12},clip]{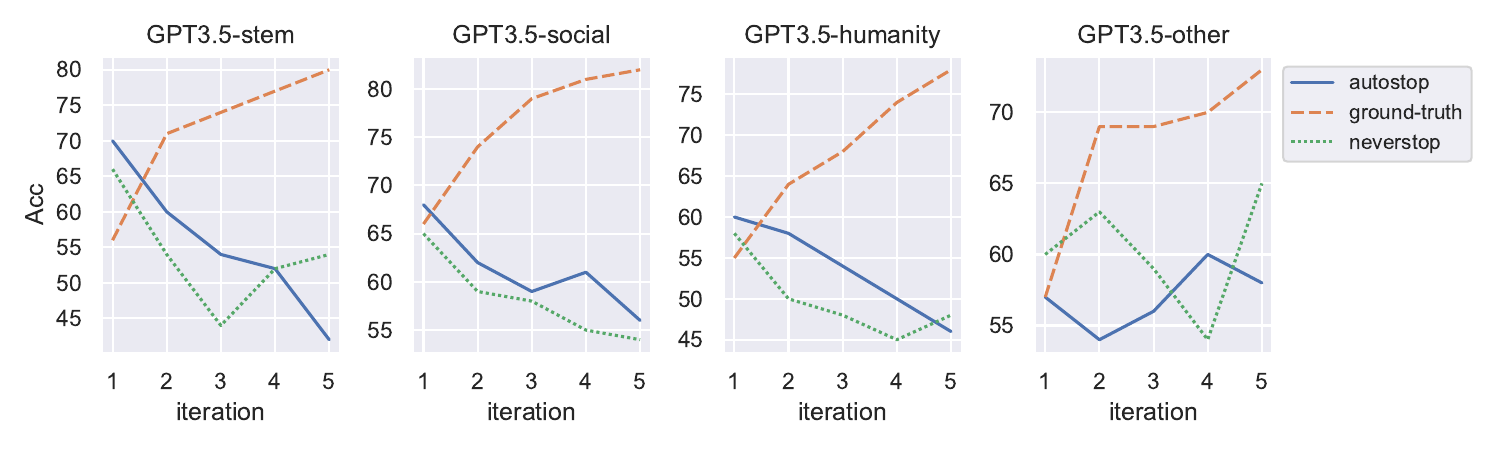}
    \vspace{-2mm}
    \caption{\footnotesize Without ground truth for validating LLM-generated outputs, LLMs struggle to consistently improve their own outputs due to their incapability of self-assessment. \textsl{Autostop} and \textsl{Neverstop} provide different generic feedback without leaking the correctness of the current response.}
    \label{fig:fail_self_asses}
\end{figure*}

The inefficiency of self-assessment and feedback generation capabilities largely hinders the performance of iterative refinements. On one hand, as depicted in  Figure~\ref{fig:fail_self_asses}, it is evident that in the absence of a ground truth reference, LLMs fail to consistently improve their predictions, indicating their limitations in self-assessment\footnote{Details of \textsl{Autostop} and \textsl{Neverstop} are in Appendix~\ref{app:Autostop}.}. On the other hand, even when ground truth labels are available, LLMs often fail to adhere to instructions for revising their incorrect predictions, as shown in Figure \ref{fig:unchanges_all}. Each bar represents the number (averaged over 5 iterations) of revised (blue) and unchanged samples (grey) among the incorrectly predicted samples. It is undesirable to see that a large number of incorrect predictions stay unchanged, 
suggesting that LLMs can become trapped in a reasoning loop.

To address the aforementioned limitations and generate high-quality feedback without relying on human experts, we propose a novel framework, refer to as \texttt{Mirror} (\textbf{M}ult\textbf{i}ple-pe\textbf{r}spective self-\textbf{r}eflection method for kn\textbf{o}wledge-rich \textbf{r}easoning). \texttt{Mirror} enables LLMs to reflect from multiple-perspective clues and 
this is achieved in a heuristic manner between a Navigator and a Reasoner, resembling a typical human tutoring process. 
For example, when tackling a complex scientific problem, 
the Navigator generates clues of key elements and rationales behind posing the question, which are crucial in 
focusing the response on the essential aspects. This information,  tailored to the question, serve as instructions for prompting the Reasoner to adjust their predictions accordingly and avoid getting stuck at a particular stage.

To initiate the unsupervised self-reflection properly and avoid being trapped in one reasoning loop, \texttt{Mirror} integrates an intrinsically motivated planning algorithm to search for the optimal reasoning trajectory. Inspired by the findings in \S\ref{sec:kg_gd} and \S\ref{sec:align_inst}, we propose to reward both the diversity of generated directions and the agreement among strategically induced perturbations in responses. Notably differing from existing tree-based planning methods for reasoning~\citep{DBLP:journals/corr/abs-2305-14992,DBLP:journals/corr/abs-2310-04406}, \texttt{Mirror} avoids deteriorated searching space by encouraging diverse generative outcomes from LLMs at each reflection step, and enhances the self-assessment ability by considering the agreements among multiple-perspective perturbations strategically induced in responses. We evaluate the performance of \texttt{Mirror} on two categories of reasoning tasks: MMLU~\citep{hendryckstest2021}, a knowledge-rich question-answering dataset, and FEVER~\citep{Thorne18Fever}, a fact-checking dataset. \texttt{Mirror} achieves a significant average improvement of over 15\% compared to recent popular unsupervised self-refinement methods. The empirical observations demonstrate that the proposed diversity-based reward and answer assessment strategy serve as reliable sources for performance enhancement.

\vspace{-1mm}
\section{Related Work}
\vspace{-1mm}

\paragraph{Self-Reflection LLMs.}
Extensive research~\citep{honovich-etal-2022-true-evaluating,Xie2023SelfEvaluationGB} has been conducted to enhance LLMs through the concept of self-reflection, where LLMs
learn from automatically generated feedback to understand and reflect on their own outputs. This feedback can stem from various sources: the LLM itself~\citep{DBLP:journals/corr/abs-2303-17651,Shinn2023ReflexionLA}, a separately trained critic module~\citep{Gou2023CRITICLL,DBLP:journals/corr/abs-2302-12813} or external sources~\citep{DBLP:conf/iclr/YaoZYDSN023}, such as Wikipedia or an Internet Browser. \citet{Gou2023CRITICLL,DBLP:journals/corr/abs-2302-12813} argued that evaluators trained on task-oriented feedback offer superior performance. For example, Refiner~\cite{Paul2023REFINERRF} took context and hypotheses as input to generate templates-based feedback for various error types. Recent studies~\citep{DBLP:journals/corr/abs-2302-12813,Shinn2023ReflexionLA,DBLP:journals/corr/abs-2305-14992} have fully utilized the in-context learning capability of LLMs, prompting them to generate high-quality feedback based on their previous generation or potential templates. \citet{DBLP:journals/corr/abs-2303-17651} proposed multiple task-oriented metrics and prompted LLMs to evaluate their own outputs based on these criteria. Similarly,~\citet{DBLP:journals/corr/abs-2302-12813,Glaese2022ImprovingAO} adopted external tools to predict multi-facet human preference scores. Our solution aligns with this trend by aiming to provide informative and customized instructions tailored to the specific task and query. Moreover, it seeks to achieve this without relying on human intervention or external tools, thereby rendering self-refinement more feasible in practice.

\paragraph{Reasoning models augmented with tree search.}
Recently, tree-based reasoning has attracted significant attention, such as Tree-of-Thought (ToT)~\citep{Yao2023TreeOT}, Grace~\citep{Khalifa2023GRACEDC}, and SelfEval-Decoding~\citep{Xie2023SelfEvaluationGB}. At each reasoning step, ToT adopts breadth-first search and depth-first search, while the latter two methods select the top-$k$ scoring candidates during the decoding process. Moreover, Monte-Carlo Tree Search (MCTS) is one of the popular search algorithms~\citep{DBLP:journals/air/SwiechowskiGSM23}, which strikes a balance between exploitation and exploration. Some existing approaches establish a reinforcement learning framework to maximize reward through learning optimal actions/states~\citep{DBLP:journals/corr/abs-2302-00111,DBLP:journals/corr/abs-2305-16209,zhu-etal-2023-solving}. Other studies fully utilize the capability of LLMs for interaction and feedback generation. For instance, RAP~\citep{DBLP:journals/corr/abs-2305-14992} leveraged step-wise rewards from interactions with the world model to decompose and solve the problem step-by -step, rather than a iterative manner. LATS~\citep{DBLP:journals/corr/abs-2310-04406} was the first work in leveraging MCTS for self-reflection. However, their feedback contains information from comparisons with ground truth, which is not applicable in our case. 
Instead, our approach, \texttt{Mirror} has no access to gold labels, and we incorporate a novel diversity reward to avoid the inefficient search in the reflection iteration.  

\section{Lost in the Reasoning Loop}
\label{sec:alignment}
Given the observed challenges in enhancing LLMs' self-improvement without  ground truth labels, particularly in knowledge-rich reasoning tasks, our initial experiment aims to address these challenges by breaking them down into two sub-questions.
\begin{itemize}[leftmargin=-2pt,itemsep=-3pt,topsep=2pt]
    \item[] $\mathcal{Q}1$: \textit{To what extent can LLMs  assess the correctness of a statement?} This investigation involves enhancing their capabilities through supervised training.
    The primary goal is to discern if there are viable solutions to enhance the verification ability of LLMs on knowledge-rich statements. 
    \item[] $\mathcal{Q}2$: \textit{How well can LLMs generate high-quality feedback to guide their own subsequent response update?} It is especially challenging when the feedback generation models are not trained on high-quality data, relying solely on the in-context learning capability of LLMs. 
\end{itemize}

\subsection{LLMs in Knowledge Grounding}
\label{sec:kg_gd}
We experiment with the multiple-choice dataset, MMLU~\cite{hendryckstest2021}, covering 57 subjects across STEM, Humanity, Social and other domains. To evaluate the ability of LLMs in assessing the knowledge-rich statements, we construct the positive and negative statements by substituting the question with the correct choice and a randomly selected choice from the other three incorrect choices, respectively. Table~\ref{tab:assess_correctness} presents the assessment accuracy of assessing. There are three categories of methods: in-context learning, fine-tuned on statements, and classification based on intermediate activations from LLMs. 

As illustrated in the first group results in Table~\ref{tab:acc_consistency}, an increase in accuracy is observed as the size of Llama-2-13B-chat increases. Notably, GPT-3.5 with 175B parameters consistently achieves the best results across the three domains, although the improvement is not directly proportional to the parameter size. We then apply advanced prompting techniques, i.e., UniLangCheck~\citep{Zhang2023InterpretableUL} on the best-performing method, GPT-3.5. Our analysis reveals that the improvements are predominantly driven by self-consistency, while UniLangCheck does not consistently contribute to improvement in grounding. For UniLangCheck, we firstly prompt LLMs to generate a fact about the key elements in a question before making the final assessment. It can be partially explained by the accumulation error, i.e., the inaccurate facts generated by LLMs before reaching the final conclusion can affect the outcome. We also calculate the correlation between accuracy and self-consistency, represented by the probability of generating a single answer through multiple repeated prompting. The average correlation $R^2$ for questions in the MMLU datasets across three LLMs is about 0.85, indicating that self-consistency can be relied upon as a proxy for assessment~\footnote{Experiment details are shown in Appendix~\ref{app:kg_ex}, self-consistency evaluation results are shown in Table~\ref{tab:acc_consistency}.}. 

We also evaluate the performance of some \textit{supervised methods} (denoted with $\star$ in Table~\ref{tab:assess_correctness}). TRUE~\citep{honovich-etal-2022-true-evaluating} involves fine-tuning a T5~\cite{JMLR:v21:20-074} model on a collection of
natural language inference (NLI) datasets for fact-checking. We further fine-tune its classifier head on our training set. ActivationRegress~\citep{Marks2023TheGO} trains classifiers using activations extracted from Llama2-13B 12-layer encodings as inputs. ContrastSearch~\citep{DBLP:conf/iclr/BurnsYKS23} is trained using contrastive and consistency loss while having no access to the factual labels. This is achieved by constructing data pairs that include both a positive-labeled and negative-labeled statements, irrespective of the true factual labels. It is surprising that both TRUE and ActivationRegress are inferior than the unsupervised ContrastSearch. 
\begin{table}[t]
\centering
\resizebox{0.45\textwidth}{!}{%
\begin{tabular}{lrrr}
\toprule[1pt]
Model  & STEM & Social & Humanity \\
\hline
Llama-2-13B-chat & 0.541&0.540&0.525\\
Llama2-70B-chat & 0.569 & 0.593 & 0.587\\
Vicuna-v1.5-13B & 0.539&0.580&0.558\\
GPT-3.5(175B) & 0.666 &0.725 &0.733\\
:+UniLangCheck &0.621 & 0.729&0.713  \\
:+Self-Consistency & 0.712&0.730 & 0.752\\
\hline
TRUE$^{\star}$& 0.545 & 0.532 &0.559 \\
\hline
ActivationRegress$^{\star}$ & 0.531 & 0.529& 0.553\\
ContrastSearch& 0.606 & 0.645 & 0.617\\
\bottomrule[1pt]
    \end{tabular}
    }
    \vspace{-2mm}
    \caption{\footnotesize The (binary classification) accuracy in evaluating the factual correctness of statements in the MNLU dataset. Methods denoted with $\star$ can access to fact labels.}
    \label{tab:assess_correctness}
\end{table}
\vspace{-2mm}

\begin{figure*}[t]
\centering
\includegraphics[width=0.89\textwidth,trim={0 14 0 0},clip]{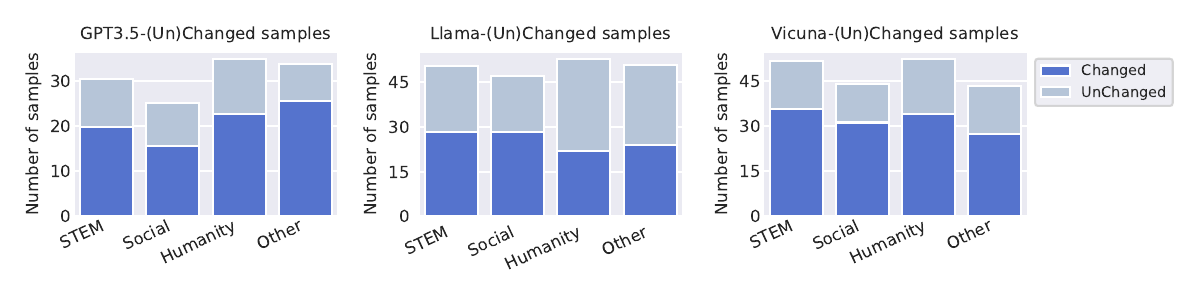}
\caption{\footnotesize The average number (across all iterations) of changed and unchanged samples among those predicted incorrectly. Large percentage of unchanged samples indicate the limited capability for efficient reflection.}
\label{fig:unchanges_all}
\end{figure*}


\subsection{LLMs in Feedback Generation}
\label{sec:align_inst}
Evaluating the quality of generated feedback poses a significant challenge, particularly when such feedback is utilized across diverse tasks~\citep{DBLP:journals/corr/abs-2303-17651}. Drawing inspiration from the pivotal role of feedback in the self-improvement, we propose to leverage the performance of LLMs in subsequent iterations for evaluation. Specifically, LLMs can access to ground truth, enabling them to evaluate the correctness of their current responses. This information is then integrated into feedback generation. Consequently, we assess the quality of feedback by examining the percentage of examples that are incorrectly answered, along with the percentage of instances where responses in the next round are revised for the same incorrectly answered examples. This comparison sheds light on the effectiveness of instructions in guiding LLMs to rectify their erroneous responses. Firstly, we follow the settings in~\citep{Shinn2023ReflexionLA} to incorporate the assessment results in the feedback: \texttt{"Observation: The answer is incorrect."} is inserted after presenting the question and previous attempt, and the LLMs are required to generate refection and response to this question again. From the results in Figure~\ref{fig:unchanges_all}, it is consistently observed across different model scales that LLMs struggle to update their predictions despite receiving explicit negative feedback. The average percentage of successfully updated examples for GPT-3.5, Llama, and Vicuna are 65.6\%, 51.79\% and 74.09\%, respectively, indicating an ample room for improvement. 

Motivated by the following two observations: (1) LLMs are particularly susceptible to context influence at the beginning or near the end~\cite{DBLP:journals/corr/abs-2307-03172}, (2) In-Context Learning is highly sensitive to stylistic and emotional words in  demonstrations~\cite{DBLP:conf/emnlp/MinLHALHZ22,li2023emotionprompt}, we develop three prompting strategies for feedback generation. An incorrectly predicted example with different prompting strategies is shown in Figure~\ref{fig:alignment_prompts}. The results in Table~\ref{tab:ans_changes_after} and Table~\ref{tab:follow_inst} suggest that based on correct question assessment, enhancing the exploration capability within a diverse answer space could lead to higher accuracy in answering knowledge-rich questions. 

The above empirical findings regarding the two research questions provide valuable insights for our proposed model, named \texttt{Mirror}. Distinguishing itself from existing self-improvement methods, \texttt{Mirror} makes two significant contributions: (1) it features a Navigator module for generating multiple question-adaptive directions, with diversity constraints implemented to prevent invalid reflections. (2) \outline{it relies on the consistency of the inherent multiple perspectives for boosted self-assessment}.

\vspace{-2mm}
\section{The Framework of Mirror}
\vspace{-2mm}
\label{sec:mcts}

\begin{figure}[h]
    \centering
    \includegraphics[trim={10 450 520 10},clip,width=0.49\textwidth]{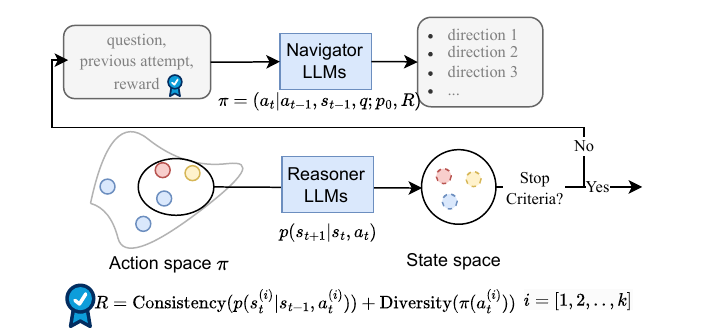}
    \caption{\footnotesize An overview of \texttt{Mirror}. It facilitates diverse question-specific directions (represented by different colored dots in the action space) to encourage extensive reflection by the Reasoner. The stopping criterion is based on the consistency among states from multiple perspectives, which also contributes to the direction generation. 
    }
    \label{fig:mcts_overview}
\end{figure}

In this section, we introduce our unsupervised self-reflection framework, \texttt{Mirror}, depicted in Figure~\ref{fig:mcts_overview}. The reward $\mathcal{R}$ consists of Diversity and Consistency terms. Diversity is applied to prevent reflection from becoming stuck and to facilitate intra-consistency involved in the stop criteria for self-assessment. The Consistency reward also influences direction generation.


\subsection{Problem Setup}
Given a question, the Reasoner is to arrive at the final answer through interacting with a Navigator. We consider a Markov Decision Process (MDP) defined by a tuple $(\mathcal{S},\mathcal{A},\mathcal{P}, \mathcal{\pi}, \gamma, \mathcal{R})$, where the $s_{t} \in \mathcal{S}$ and $a_{t} \in \mathcal{A}$ denote the state and action, respectively in the $t$-th reflection iteration. In the context of multiple-choice question, $a_{t}$ is the direction generated by the Navigator, and $s_{t}$ is the response generated by the Reasoner, including the answer to the question and the rationale behind. $\mathcal{R}(s,a)$ is the reward function. Therefore, we have state transition distribution $\mathcal{P}(s_{t}|s_{t-1},a_{t-1})$ and action generation distribution $\mathcal{\pi}(a_{t}|s_{t-1},a_{t-1},q,p_0,\mathcal{R})$, where $p_0$ is the prompt for the Navigator to generate direction $a_{t}$. It is nontrivial to obtain frequent rewards that incentivize self-refinement progress without access to the ground truth. Therefore, we turn to an intrinsically motivated planning algorithm, i.e., Monte-Carlo Tree Search (MCTS)~\citep{kocsis2006bandit,6145622,DBLP:journals/air/SwiechowskiGSM23} to efficiently explore the environment augmenting rewards with auxiliary objectives~\citep{mu2022improving,DBLP:conf/icml/DuWWCDA0A23}.

Comparing to existing work search-based reasoning methods based on frozen LLMs~\citep{DBLP:journals/corr/abs-2305-14992,DBLP:journals/corr/abs-2310-04406}, we highlight two notable contributions addressing the vulnerabilities of LLMs as discussed in \S \ref{sec:alignment}: (1) \textsl{Step-wise Multiple-perspective self-assessment:} unlike approaches that rely on ground truth or majority-voting based on several complete generated trajectories, our framework utilizes multiple-perspective consistency as stop criteria at each step $t$. (2) \textsl{Novel Reward Mechanism:}  a novel diversity mechanism is designed to avoid the null space encountered in traditional random search settings. Our method is detailed in Algorithm~\ref{alg:cap} in the Appendix. 

\subsection{Multiple-perspective Assessment}
\label{sec:answer_assess}
Motivated by the empirical results in \S~\ref{sec:kg_gd} regarding knowledge-grounding, we propose to employ an advanced consistency-based method as a surrogate for factual correctness when external resources are unavailable.  This method considers both intra- and inter-consistency of the generated responses. Specially, we employ the Navigator for $K$ question-oriented direction generation, $a \sim \pi(a_{t}|s_{t-1},a_{t-1},q,p_{0},\mathcal{R})$.
These $K$ directions are intended to provide diverse perspectives for problem-solving, with the agreement among guided responses representing inter-consistency. Meanwhile, the confidence in self-consistency~\citep{DBLP:conf/iclr/0002WSLCNCZ23} serves as the measure of intra-consistency. 

To integrate consistency considerations into the assessment per reflection iteration, we use \textbf{intra-consistency} to determine whether the Reasoner should accept its initial response. If the intra-consistency for our initial answer surpasses a threshold $T_0$, we consider it as the final result; otherwise, we integrate the \textbf{inter-consistency} as an indicator for stopping criteria in subsequent reflection iterations. We derive the final answer when the inter-consistency exceeds $T_0$ or when reach the predefined maximum iterations, selecting the final answer with the highest consistency score~\footnote{The threshold $T_0$ for different models are datasets are set according to the validation performance, details in~\ref{app:inplement_details}.}. This inter-consistency also becomes part of reward $\mathcal{R}_\text{consistency}$ for the current state and contribute to the direction generation. Besides, the intra-consistency value is transformed into verbal form, becoming part of the prompt $p_{0}$ given for Navigator to generate direction. This is inspired by our observation that higher intra-consistency implies a higher likelihood of correctness, so we offer this additional information to assist in feedback generation. This is similar to~\citep{DBLP:journals/corr/abs-2305-08848}, where ICL performance benefits from accessing to the prediction of a supervised fine-tuned smaller model. Our assessment method is different from majority vote, which treats every node with the same weight when aggregating the final result. The comparison results are shown in Table~\ref{tab:answer_asses}. 

\subsection{Diverse and Valid Search Space}
Obtaining a meaningful and diverse action space is challenging due to the absence of a dense and well-defined reward function in the planning algorithm. One of the predominant reasons is that different action sequences can lead to similar outcomes~\citep{DBLP:journals/ras/BaranesO13}. In our context, considering the limitation of LLMs in following instructions, the Reasoner may ignore the differences among multiple directions and generate identical responses merely based on the question. Therefore, some intrinsically motivated reinforcement learning algorithms choose to explore outcomes rather than actions~\citep{oudeyer2007intrinsic,ladosz2022exploration}. MCTS addresses the limitation of sparse rewards by visiting novel states or transitions through random exploration~\citep{DBLP:conf/icml/DuWWCDA0A23}. The most popular algorithm in the
MCTS family, Upper Confidence Bound for Trees (UCT)~\citep{kocsis2006bandit} is treated as the choice of child node, $\text{UCT} = \overline{\mathcal{R}}_{j}+2C_{p}\sqrt{\frac{2\,\text{In}\, N(n)}{N(n_j)}}$, where $\overline{\mathcal{R}}_{j}$ is the average reward for child node $j$, while the second term encourages sampling from nodes whose children are less visited.
$N(n)$ is the number of times current node (parent) has been visited in previous iterations, and $N(n_j)$ is times of the child node has been visited. The $C_p > 0$ is a constant to control balance between exploitation (first term) and exploration (second term).
In our case, we specifically promote diversity between the parent and child node, i.e., the response in previous attempt $s_{t-1}$ and the current attempt $s_{t}$. For multiple-choice questions in MMLU,  we assess if the predicted choices are the same across two reflection iterations. The discrepancy in responses indicates the alleviation of null direction space and the avoidance of being stuck, especially given the relatively low consistency with the response from the previous iteration. The relationship between task performance and the diversity of responses in the generated tree, as illustrated in Figure~\ref{fig:acc_pres}, confirms our motivation for diversity enhancement.

However, maximizing diversity of outcomes may not always be enough, as less relevant states might be collected~\citep{DBLP:conf/icml/DuWWCDA0A23}. Therefore, to ensure valid search space, we filter out states whose associated responses are not in the correct form, such as failing to provide a final choice, or refusing to answer questions for moral considerations. For search efficiency, our proposed stopping criteria is to terminate the search process once its inter-consistency surpasses a threshold, thereby avoiding unnecessary search and expansion costs.

\section{Can Mirror Steer LLMs in Iterative Improvements?}
\label{sec:mcts_ex}
We evaluate our proposed \texttt{Mirror} on MMLU and  FEVER~\citep{Thorne18Fever}. FEVER is a fact-checking dataset featuring three labels for knowledge-rich statements, i.e., \texttt{supports}, \texttt{refutes} and \texttt{not enough info}.\footnote{We also include the evaluation results on GSM8K~\citep{DBLP:journals/corr/abs-2110-14168} in Table~\ref{tab:gpt35_math}.}

\subsection{Experimental Setup and Results}
\paragraph{Comparison methods.} The evaluation models are GPT-3.5, Llama2-13B-Chat~\citep{Touvron2023Llama2O}, and Vicuna-v1.5-7B~\citep{Zheng2023JudgingLW}~\footnote{We denote them as Llama13B and Vicuna13B for simplicity. Experiment details can be found in Appendix~\ref{app:mcts_ex}.}. We equip the LLMs with different reasoning mechanisms, including Chain-of-Thought (CoT)~\citep{wei2022chain}, Self-consistency~\citep{DBLP:conf/iclr/0002WSLCNCZ23}, Self-Correction~\citep{DBLP:journals/corr/abs-2310-01798} and Reflexion(w.GT)~\cite{Shinn2023ReflexionLA}. We implement CoT by prompting LLMs to first generate step-by-step thoughts and then generate answers based on those thoughts. We repeat this process for five times, resulting in Self-Consistency$^{(5)}$. The remaining two methods are self-improvement techniques where LLMs are first prompted to generate reflections, followed by updating their current response accordingly if applicable. Self-Correction relies on LLM's internal knowledge for answer assessment, while Reflexion compares the current answer with the ground truth for evaluation. 
\begin{figure*}[t]
    \centering
    \includegraphics[trim=0 235 90 15,clip,width=0.99\textwidth]{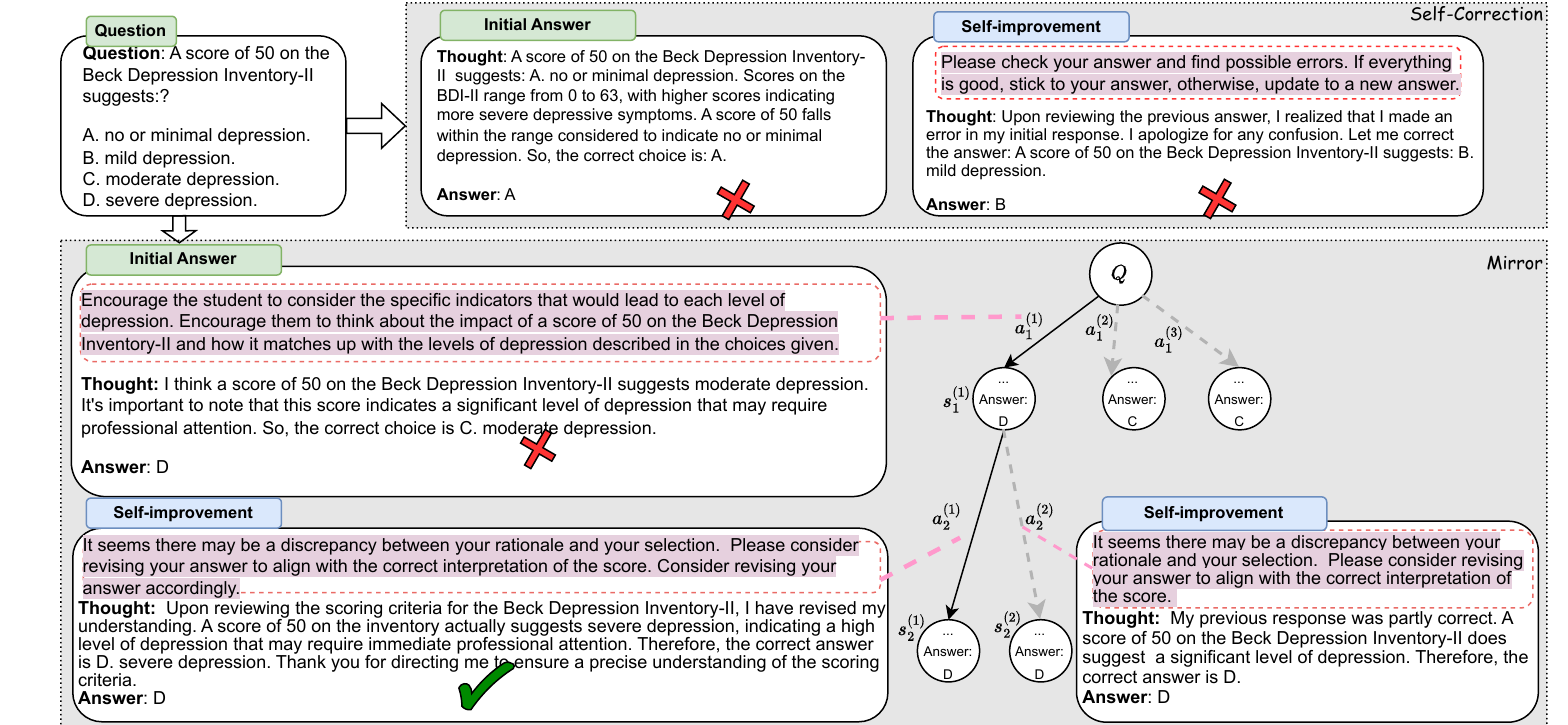}
    \caption{\footnotesize Reasoning process of self-correction and \texttt{Mirror}. Text in red are generated directions. Our diversity is characterised in (i) generating directions tailored to questions (ii) encouraging exploration in multiple plausible reasoning paths. The final answer is derived through an agreement among multiple trajectories.}
    \label{fig:tree_example}
\vspace{-5pt}
\end{figure*}


\begin{table}[h]
    \centering
    \resizebox{0.489\textwidth}{!}{%
    \begin{tabular}{l|rrrrr}
    \toprule[1pt]
    \textbf{Methods} & \textbf{STEM}&\textbf{Social}&\textbf{Hum}&\textbf{Others}&\textbf{FEVER}\\
    \midrule
    Relexion(w.GT)$^{(5)}$&0.79&0.84&0.78&0.73&0.72\\
     GPT-3.5 (CoT)& 0.63& 0.65& 0.53&0.60&0.58\\
     Self-Consistency$^{(5)}$ &0.67&0.68&0.58&0.64&0.61 \\
     Self-Correct$^{(2)}$&0.63&0.62&0.55&0.54&0.55\\
     \texttt{Mirror} & 0.76&0.77 & 0.71&0.67&0.64\\
\midrule
    Relexion(w.GT)$^{(5)}$&0.64&0.63&0.60&0.64&0.59\\
     Llama13B(CoT)& 0.42 & 0.58&0.42&0.53&0.40  \\
     Self-Consistency$^{(5)}$&0.45&0.60&0.49&0.57&0.46 \\
     Self-Correct$^{(2)}$& 0.42&0.52&0.53&0.45&0.36 \\
    \texttt{Mirror} & 0.57&0.62&0.58 &0.62&0.54\\
    \midrule
         Relexion(w.GT)$^{(5)}$&0.62& 0.68&0.59&0.69&0.59\\
     Vicuna13B (CoT)&0.46&0.57&0.43&0.57&0.39  \\
     Self-Consistency$^{(5)}$&0.50&0.62&0.53&0.60&0.43 \\
     Self-Correct$^{(2)}$&0.43&0.49&0.42&0.49&0.38 \\
    \texttt{Mirror} & 0.59&0.64&0.56&0.65&0.46\\
    \bottomrule[1pt]
    \end{tabular}
    }
    \caption{\footnotesize Performances of different reasoning methods, with an upper-bound represented by results obtained when ground truth is provided, denoted as Relexion(w.GT). The superscripts denote the number of reasoning iterations.}
    \label{tab:main_results}
\end{table}

\paragraph{Results.}
\label{sec:inst_ex}
The results are shown in Table~\ref{tab:main_results}. 
By comparing CoT with Self-Correction, we observe the performance degradation after two rounds of self-Correction across almost all datasets and models. This observation aligns with our findings in \S\ref{sec:kg_gd} and in~\cite{DBLP:journals/corr/abs-2310-01798}. Equipped with self-consistency$^{(5)}$, significant performance improvements are evident across all settings. \texttt{Mirror} considers additional inter-consistency, achieves the most notable improvements, with a relative increase of more than 15\% across the three models. Figure~\ref{fig:tree_example} illustrates the reasoning process of Self-correction and \texttt{Mirror}. Both methods fail to answer correctly in the first trial. With question-oriented direction, the Reasoner better identify errors in the initial response, such as, \textit{the error in score direction} and \textit{inconsistency between rationales and selection}. The consistency-based criteria built in the tree further improves the fact assessment. During backpropagation, node $s^{(1)}_{1}$ receives a higher reward, leading to the leftmost reasoning path (details of direction $a^{(1)}_{1}, a^{(1)}_{2}, a^{(2)}_{2}$ and corresponding responses are shown in the text frame). By contrast, Self-correction seems to engage in groundless inference by switching answers without explicit clues. 
Even comparing \texttt{Mirror} with Relexion(w.GT), we find comparable results for GPT-3.5 on the STEM dataset, for Llama on all datasets except for STEM and for Vicuna on STEM and Humanity. From the perspective of the model, the average improvements over baselines for GPT-3.5 are particularly prominent, partly explained by its better ability to adhere to provided directions. This can also explain the marginal improvements even ground truth are accessible to the smaller models.
\subsection{Analysis}
We discuss the effects of key strategies in \texttt{Mirror}.
\paragraph{Question-Oriented Direction.}
Motivated by the findings in \S~\ref{sec:align_inst} that LLMs struggle to effectively reflect on themselves with generic feedback, \texttt{Mirror} is equipped with a Navigator for generating question-oriented directions. To study the effects of these directions (results in Table~\ref{tab:direction_ablation}), we adopt our Navigator for direction generation for CoT settings, in which the direction (GenerativeDirect) is introduced before the LLM generates its thought on the previous trial. We then replace all adaptive directions with a single generic direction (FixedDirect) which reads: \texttt{
Read the question and choices carefully and diagnose the previous response by locating the incorrect clues and update the response if applicable.} 
Comparing with CoT, the inclusion of GenerativeDirect boosts the performance across all settings with significant improvements. Conversely, FixedDirect sometimes results in performance degradation for Llama13B. The impact of FixedDirect is similar to advanced instruction intended to provide general direction for the task, whereas GenerativeDirect offers question-specific advice to accurately summarize clues for solution. Referencing to the example in Figure~\ref{fig:inst_ex}, \texttt{Mirror} (bottom) firstly prompts the Navigator for direction generation (highlighted in red), which captures the key elements, such as \textit{``the characteristics of a connected and undirected graph''}. The Reasoner then follows this direction to explain the key concepts of this graph, laying a solid foundation for reaching the correct conclusion. Without such direction, the Reasoner may overlook or misinterpret knowledge about this graph, leading to errors in the conclusion. 
\begin{table}[ht]
    \centering
    \resizebox{0.40\textwidth}{!}{%
    \begin{tabular}{lp{3cm}|c|c}
    \toprule[1pt]
     \textbf{Models}&\textbf{Methods}  & \textbf{MMLU} & \textbf{FEVER}  \\
     \midrule
       GPT-3.5:&CoT  & 0.68 & 0.58 \\
       &+ FixedDirect & 0.73 & 0.60\\
       &+ GenerativeDirect& 0.78 &0.64\\
       \midrule
          Llama13B:& CoT  &0.46 &0.40  \\
       &+ FixedDirect &0.43 &0.39\\
       &+ GenerativeDirect &0.49 &0.45 \\
       \midrule
        Vicuna13B:&   CoT  & 0.48  &0.42\\
       &+ FixedDirect& 0.51 & 0.43\\
       &+ GenerativeDirect&  0.55&0.45\\
    \bottomrule[1pt]
    \end{tabular}
    }
    \caption{Performances of using generic fixed direction and generative direction on top of CoT.}
    \label{tab:direction_ablation}
\end{table}

\begin{figure}[h]
    \centering
    \includegraphics[width=0.49\textwidth,trim={4 5 0 8},clip]{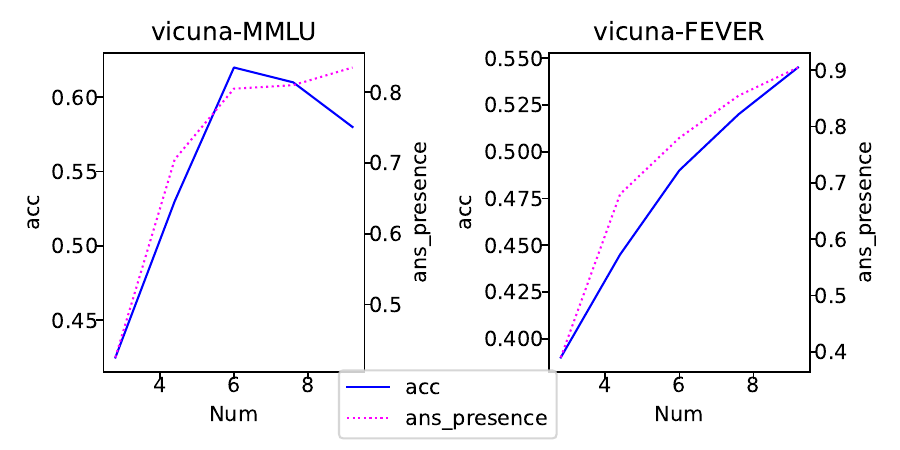}
    \caption{\footnotesize The Accuracy (acc) and the percentage of samples where the ground truth is included in the tree (ans-presence), with different sizes of search space (Num). Results for GPT-3.5 and Llama13B are in Figure~\ref{fig:acc_pres_other1} and~\ref{fig:acc_pres_other2}.}
    \label{fig:acc_pres}
\end{figure}

\paragraph{Diversity of the Search Space.}
We demonstrate the impact of multiple-perspective directions, aiming at guiding the Reasoner out of reflection traps. To this end, we compute the percentage of generated trajectories containing the correct answers ($\text{ans}_{\_}\text{presence}$) and the according task performances (acc) across various action space sizes, i.e., the number of generated directions. The results in Figure~\ref{fig:acc_pres} indicate that lager search space enhanced by the $\mathcal{R}_\text{diversity}$ can increase the probability of reaching the correct answer. 

We provide a case to show the effects of diversity in directions (highlighted difference are in bold). The quantitative results of semantic similarity between multiple directions are shown in Table~\ref{tab:diversity}.

\begin{myexamplea}
\small \textbf{\textcolor{blue!90!black}{Question}}: The price elasticity of demand for a product is greater if
\tcbline\vspace{-3mm}
\textbf{\textcolor{blue!90!black}{- Answer without Direction}}: \\
\emph{C. the number of substitute products is limited.}
\tcbline\vspace{-3mm}
\textbf{\textcolor{blue!90!black}{- Answer with Direction}}:
\vspace{2mm}
\\
\textbf{\textcolor{blue!90!black}{Direction 1}}: Pay close attention to nuances among options and prioritize selections that resonate with \textbf{economic principles} and the factors influencing consumer behavior.
\begin{itemize}[leftmargin=*,itemsep=0pt,topsep=2pt,]
\item[] \textbf{\textcolor{blue!90!black}{ Answer 1}}: Given the advice, the most suitable choice would be: \emph{A. the proportion of the good of the consumer's budget is high.} This aligns with \textbf{economic principles of demand elasticity}.
\end{itemize}
\textbf{\textcolor{blue!90!black}{Direction 2}}: Prioritize options that align with factors impacting consumer responsiveness to price changes, such as \textbf{budget allocation} and \textbf{substitute availability}, while considering the time frame for consumer reactions.
\begin{itemize}[leftmargin=*,itemsep=-2pt,topsep=1pt,]
    \item[] \textbf{\textcolor{blue!90!black}{Answer 2}}: Given the advice, the most suitable choice would be: \emph{A. the proportion of the good of the consumer's budget is high.} This choice suggests that when a significant portion of \textbf{a consumer's budget} is allocated to a particular product, they are likely to be more sensitive to price changes.
\end{itemize}
\end{myexamplea}

\paragraph{Performance of Answer Assessment Criteria.} As discussed in Section~\ref{sec:kg_gd}, LLMs struggle to assess the correctness of knowledge-rich statements, a capability that can be consistently enhanced through self-consistency. We further reform the majority-voting assessment process by considering the inter-consistency built in the hierarchical decision-making tree. To study the effects of our answer assessment criteria described in \S\ref{sec:answer_assess}, we compare them with two other voting methods, i.e., self-consistency and majority vote within our generated tree-trajectories. We average the results from Table~\ref{tab:main_results} for CoT and Self-consistency$^{(5)}$ across four domains in MMLU and denote them as CoT$^{(1)}$ and CoT$^{(5)}$, respectively. We also compare with CoT$^{(15)}$ because our generated trees have at most 3 layers and 5 branches in each layer, so there are 15 candidate nodes. For Majority$^{(tree)}$, we select the final answer through majority-voting among all intermediate nodes in our generated tree-trajectories. The results of different final answer assessments are presented in Table~\ref{tab:answer_asses}. The performance improvements are observed on self-consistency$^{(15)}$ over self-consistency$^{(5)}$, although the improvements percentage isn't pronounced as that seen in the comparison between self-consistency$^{(5)}$ over CoT. We observe a performance increase after applying majority-voting in the CoT settings, while this simple strategy doesn't yield improvements in the generated tree. This is because undesirable responses may be generated during the node expanding phase, and majority voting treats all nodes equally. In contrast, our reward-based search tends to focus on reliable nodes with higher confidence in each reflection step, thereby avoiding search cost on less desirable nodes. 

\begin{table}[ht]
    \centering
    \resizebox{0.42\textwidth}{!}{%
    \begin{tabular}{lrcc}
    \toprule[1pt]
    \textbf{Models}&\textbf{Ans. Assessment} & \textbf{MMLU} & \textbf{FEVER} \\
    \midrule
  GPT-3.5:&CoT$^{(1)}$&0.60&0.58\\
  & CoT$^{(5)}$&0.64&0.61 \\
   & CoT$^{(15)}$&0.67&0.62 \\
  &+Majority$^{(tree)}$ &0.69&0.59  \\
    &+Reward Search$^{(tree)}$   &0.73&0.64 \\
    \midrule
  Llama13B:&CoT$^{(1)}$&0.49&0.40\\
  &CoT$^{(5)}$&0.53&0.46 \\
   & CoT$^{(15)}$&0.55&0.48 \\
  &+Majority$^{(tree)}$&0.58&0.50  \\
    &+Reward Search$^{(tree)}$&0.60&0.54 \\
    
    \midrule
 Vicuna13B:&CoT$^{(1)}$&0.51&0.39\\
 & CoT$^{(5)}$ &0.56&0.43\\
  & CoT$^{(15)}$&0.58&0.43 \\
  &+Majority$^{(tree)}$&0.59&0.43  \\
    &+Reward Search$^{(tree)}$ &0.60&0.46 \\
    \bottomrule[1pt]
    \end{tabular}
    }
    \vspace{-8pt}
    \caption{\footnotesize Results of different answer assessment methods. }
    \label{tab:answer_asses}
    \vspace{-4pt}
\end{table}

\paragraph{Search Efficiency Analysis.} Mirror is a tree-like searching algorithm that benefits from iterative reasoning process, which enhances the reasoning ability at the computation cost. To mitigate the search cost, we (i) incorporate the Monte-Carlo tree search for its selective search and expansion. (ii) introduce early-stop criteria to encourage a shallow and avoid multiple playouts. We summarise the tree-depth in Table~\ref{tab:tree_depth}. The results below show that our resulting tree, with a maximum depth of 3, is heavily unbalanced and shallow.
\begin{table}[h]
    \centering
    \resizebox{0.48\textwidth}{!}{%
    \begin{tabular}{cccccc}
    \toprule
    &\textbf{STEM}& \textbf{Social}&\textbf{Hum}&\textbf{Others}&\textbf{FEVER}       \\
    \midrule
     Depth=2   & 0.17 &0.12&0.28&0.13&0.04\\
     Depth=3 & 0.00&0.09&0.20&0.04&0.00 \\
     \bottomrule
    \end{tabular}
    }
    \caption{\footnotesize Depth of the search tree based on GPT-3.5 across different datasets.}
    \label{tab:tree_depth}
\end{table}

\section{Conclusion}
In this paper, we present a multiple-perspective reflection method, called \texttt{Mirror}, for knowledge-enriched reasoning. To tackle the limitations of LLMs in fact assessment and the generation of high-quality feedback, \texttt{Mirror} is equipped with a directional Navigator, enabling the Reasoner to identify multiple key clues in problem-solving. Furthermore, the consistency among responses generated under different directions enhances the validity of answer assessment, particularly when ground truth is not accessible. Experiments conducted demonstrate \texttt{Mirror}'s superiority over several contemporary CoT-based and self-consistency-based reasoning approaches without access to ground truth. Moreover, the ablation study results clearly show that our strategies effectively alleviate the aforementioned challenges.
\section*{Limitations}
In this study, our primary focus is to identify optimal reasoning trajectories based on generated outputs and frozen states. However, the ability to assess facts and generate reflections may be limited by the unaltered decoding process and pre-training. To fully leverage the potential of LLMs in complex reasoning, it is beneficial to explore two directions: (1) Strategically guiding fine-grained generation, such as token-level generation during the decoding phase within the expansive generation space.
(2) Fine-tuning LLMs through access to limited task-oriented data to enhance their responses to more complex problems.

\section*{Acknowledgements}
This work was supported in part by the UK Engineering and Physical Sciences Research Council (EPSRC) through a Turing AI Fellowship (grant no. EP/V020579/1, EP/V020579/2) and a New Horizons grant (grant no. EP/X019063/1).

\bibliography{custom}

\begin{thebibliography}{49}
\expandafter\ifx\csname natexlab\endcsname\relax\def\natexlab#1{#1}\fi

\bibitem[{Baranes and Oudeyer(2013)}]{DBLP:journals/ras/BaranesO13}
Adrien Baranes and Pierre{-}Yves Oudeyer. 2013.
\newblock \href {https://doi.org/10.1016/J.ROBOT.2012.05.008} {Active learning of inverse models with intrinsically motivated goal exploration in robots}.
\newblock \emph{Robotics Auton. Syst.}, 61(1):49--73.

\bibitem[{Browne et~al.(2012)Browne, Powley, Whitehouse, Lucas, Cowling, Rohlfshagen, Tavener, Perez, Samothrakis, and Colton}]{6145622}
Cameron~B. Browne, Edward Powley, Daniel Whitehouse, Simon~M. Lucas, Peter~I. Cowling, Philipp Rohlfshagen, Stephen Tavener, Diego Perez, Spyridon Samothrakis, and Simon Colton. 2012.
\newblock \href {https://doi.org/10.1109/TCIAIG.2012.2186810} {A survey of monte carlo tree search methods}.
\newblock \emph{IEEE Transactions on Computational Intelligence and AI in Games}, 4(1):1--43.

\bibitem[{Burns et~al.(2023)Burns, Ye, Klein, and Steinhardt}]{DBLP:conf/iclr/BurnsYKS23}
Collin Burns, Haotian Ye, Dan Klein, and Jacob Steinhardt. 2023.
\newblock \href {https://openreview.net/pdf?id=ETKGuby0hcs} {Discovering latent knowledge in language models without supervision}.
\newblock In \emph{The Eleventh International Conference on Learning Representations, {ICLR} 2023, Kigali, Rwanda, May 1-5, 2023}. OpenReview.net.

\bibitem[{Cobbe et~al.(2021)Cobbe, Kosaraju, Bavarian, Chen, Jun, Kaiser, Plappert, Tworek, Hilton, Nakano, Hesse, and Schulman}]{DBLP:journals/corr/abs-2110-14168}
Karl Cobbe, Vineet Kosaraju, Mohammad Bavarian, Mark Chen, Heewoo Jun, Lukasz Kaiser, Matthias Plappert, Jerry Tworek, Jacob Hilton, Reiichiro Nakano, Christopher Hesse, and John Schulman. 2021.
\newblock \href {http://arxiv.org/abs/2110.14168} {Training verifiers to solve math word problems}.
\newblock \emph{CoRR}, abs/2110.14168.

\bibitem[{Du et~al.(2023{\natexlab{a}})Du, Yang, Dai, Dai, Nachum, Tenenbaum, Schuurmans, and Abbeel}]{DBLP:journals/corr/abs-2302-00111}
Yilun Du, Mengjiao Yang, Bo~Dai, Hanjun Dai, Ofir Nachum, Joshua~B. Tenenbaum, Dale Schuurmans, and Pieter Abbeel. 2023{\natexlab{a}}.
\newblock \href {https://doi.org/10.48550/ARXIV.2302.00111} {Learning universal policies via text-guided video generation}.
\newblock \emph{CoRR}, abs/2302.00111.

\bibitem[{Du et~al.(2023{\natexlab{b}})Du, Watkins, Wang, Colas, Darrell, Abbeel, Gupta, and Andreas}]{DBLP:conf/icml/DuWWCDA0A23}
Yuqing Du, Olivia Watkins, Zihan Wang, C{\'{e}}dric Colas, Trevor Darrell, Pieter Abbeel, Abhishek Gupta, and Jacob Andreas. 2023{\natexlab{b}}.
\newblock \href {https://proceedings.mlr.press/v202/du23f.html} {Guiding pretraining in reinforcement learning with large language models}.
\newblock In \emph{International Conference on Machine Learning, {ICML} 2023, 23-29 July 2023, Honolulu, Hawaii, {USA}}, volume 202 of \emph{Proceedings of Machine Learning Research}, pages 8657--8677. {PMLR}.

\bibitem[{Gao et~al.(2023{\natexlab{a}})Gao, Dai, Pasupat, Chen, Chaganty, Fan, Zhao, Lao, Lee, Juan, and Guu}]{gao-etal-2023-rarr}
Luyu Gao, Zhuyun Dai, Panupong Pasupat, Anthony Chen, Arun~Tejasvi Chaganty, Yicheng Fan, Vincent Zhao, Ni~Lao, Hongrae Lee, Da-Cheng Juan, and Kelvin Guu. 2023{\natexlab{a}}.
\newblock \href {https://doi.org/10.18653/v1/2023.acl-long.910} {{RARR}: Researching and revising what language models say, using language models}.
\newblock In \emph{Proceedings of the 61st Annual Meeting of the Association for Computational Linguistics (Volume 1: Long Papers)}, pages 16477--16508, Toronto, Canada. Association for Computational Linguistics.

\bibitem[{Gao et~al.(2023{\natexlab{b}})Gao, Yen, Yu, and Chen}]{gao2023enabling}
Tianyu Gao, Howard Yen, Jiatong Yu, and Danqi Chen. 2023{\natexlab{b}}.
\newblock Enabling large language models to generate text with citations.
\newblock In \emph{Empirical Methods in Natural Language Processing (EMNLP)}.

\bibitem[{Glaese et~al.(2022)Glaese, McAleese, Trkebacz, Aslanides, Firoiu, Ewalds, Rauh, Weidinger, Chadwick, Thacker, Campbell-Gillingham, Uesato, Huang, Comanescu, Yang, See, Dathathri, Greig, Chen, Fritz, Elias, Green, Mokr'a, Fernando, Wu, Foley, Young, Gabriel, Isaac, Mellor, Hassabis, Kavukcuoglu, Hendricks, and Irving}]{Glaese2022ImprovingAO}
Amelia Glaese, Nathan McAleese, Maja Trkebacz, John Aslanides, Vlad Firoiu, Timo Ewalds, Maribeth Rauh, Laura Weidinger, Martin Chadwick, Phoebe Thacker, Lucy Campbell-Gillingham, Jonathan Uesato, Po-Sen Huang, Ramona Comanescu, Fan Yang, A.~See, Sumanth Dathathri, Rory Greig, Charlie Chen, Doug Fritz, Jaume~Sanchez Elias, Richard Green, Sovna Mokr'a, Nicholas Fernando, Boxi Wu, Rachel Foley, Susannah Young, Iason Gabriel, William~S. Isaac, John F.~J. Mellor, Demis Hassabis, Koray Kavukcuoglu, Lisa~Anne Hendricks, and Geoffrey Irving. 2022.
\newblock \href {https://api.semanticscholar.org/CorpusID:252596089} {Improving alignment of dialogue agents via targeted human judgements}.
\newblock \emph{ArXiv}, abs/2209.14375.

\bibitem[{Gou et~al.(2023{\natexlab{a}})Gou, Shao, Gong, Shen, Yang, Duan, and Chen}]{DBLP:journals/corr/abs-2305-11738}
Zhibin Gou, Zhihong Shao, Yeyun Gong, Yelong Shen, Yujiu Yang, Nan Duan, and Weizhu Chen. 2023{\natexlab{a}}.
\newblock \href {https://doi.org/10.48550/ARXIV.2305.11738} {{CRITIC:} large language models can self-correct with tool-interactive critiquing}.
\newblock \emph{CoRR}, abs/2305.11738.

\bibitem[{Gou et~al.(2023{\natexlab{b}})Gou, Shao, Gong, Shen, Yang, Duan, and Chen}]{Gou2023CRITICLL}
Zhibin Gou, Zhihong Shao, Yeyun Gong, Yelong Shen, Yujiu Yang, Nan Duan, and Weizhu Chen. 2023{\natexlab{b}}.
\newblock \href {https://api.semanticscholar.org/CorpusID:258823123} {Critic: Large language models can self-correct with tool-interactive critiquing}.
\newblock \emph{ArXiv}, abs/2305.11738.

\bibitem[{Hao et~al.(2023)Hao, Gu, Ma, Hong, Wang, Wang, and Hu}]{DBLP:journals/corr/abs-2305-14992}
Shibo Hao, Yi~Gu, Haodi Ma, Joshua~Jiahua Hong, Zhen Wang, Daisy~Zhe Wang, and Zhiting Hu. 2023.
\newblock \href {https://doi.org/10.48550/ARXIV.2305.14992} {Reasoning with language model is planning with world model}.
\newblock \emph{CoRR}, abs/2305.14992.

\bibitem[{Hendrycks et~al.(2021)Hendrycks, Burns, Basart, Zou, Mazeika, Song, and Steinhardt}]{hendryckstest2021}
Dan Hendrycks, Collin Burns, Steven Basart, Andy Zou, Mantas Mazeika, Dawn Song, and Jacob Steinhardt. 2021.
\newblock Measuring massive multitask language understanding.
\newblock \emph{Proceedings of the International Conference on Learning Representations (ICLR)}.

\bibitem[{Honovich et~al.(2022)Honovich, Aharoni, Herzig, Taitelbaum, Kukliansy, Cohen, Scialom, Szpektor, Hassidim, and Matias}]{honovich-etal-2022-true-evaluating}
Or~Honovich, Roee Aharoni, Jonathan Herzig, Hagai Taitelbaum, Doron Kukliansy, Vered Cohen, Thomas Scialom, Idan Szpektor, Avinatan Hassidim, and Yossi Matias. 2022.
\newblock \href {https://doi.org/10.18653/v1/2022.naacl-main.287} {{TRUE}: Re-evaluating factual consistency evaluation}.
\newblock In \emph{Proceedings of the 2022 Conference of the North American Chapter of the Association for Computational Linguistics: Human Language Technologies}, pages 3905--3920, Seattle, United States. Association for Computational Linguistics.

\bibitem[{Huang et~al.(2023)Huang, Chen, Mishra, Zheng, Yu, Song, and Zhou}]{DBLP:journals/corr/abs-2310-01798}
Jie Huang, Xinyun Chen, Swaroop Mishra, Huaixiu~Steven Zheng, Adams~Wei Yu, Xinying Song, and Denny Zhou. 2023.
\newblock \href {https://doi.org/10.48550/ARXIV.2310.01798} {Large language models cannot self-correct reasoning yet}.
\newblock \emph{CoRR}, abs/2310.01798.

\bibitem[{Kadavath et~al.(2022)Kadavath, Conerly, Askell, Henighan, Drain, Perez, Schiefer, Hatfield{-}Dodds, DasSarma, Tran{-}Johnson, Johnston, Showk, Jones, Elhage, Hume, Chen, Bai, Bowman, Fort, Ganguli, Hernandez, Jacobson, Kernion, Kravec, Lovitt, Ndousse, Olsson, Ringer, Amodei, Brown, Clark, Joseph, Mann, McCandlish, Olah, and Kaplan}]{DBLP:journals/corr/abs-2207-05221}
Saurav Kadavath, Tom Conerly, Amanda Askell, Tom Henighan, Dawn Drain, Ethan Perez, Nicholas Schiefer, Zac Hatfield{-}Dodds, Nova DasSarma, Eli Tran{-}Johnson, Scott Johnston, Sheer~El Showk, Andy Jones, Nelson Elhage, Tristan Hume, Anna Chen, Yuntao Bai, Sam Bowman, Stanislav Fort, Deep Ganguli, Danny Hernandez, Josh Jacobson, Jackson Kernion, Shauna Kravec, Liane Lovitt, Kamal Ndousse, Catherine Olsson, Sam Ringer, Dario Amodei, Tom Brown, Jack Clark, Nicholas Joseph, Ben Mann, Sam McCandlish, Chris Olah, and Jared Kaplan. 2022.
\newblock \href {https://doi.org/10.48550/ARXIV.2207.05221} {Language models (mostly) know what they know}.
\newblock \emph{CoRR}, abs/2207.05221.

\bibitem[{Khalifa et~al.(2023)Khalifa, Logeswaran, Lee, Lee, and Wang}]{Khalifa2023GRACEDC}
Muhammad Khalifa, Lajanugen Logeswaran, Moontae Lee, Ho~Hin Lee, and Lu~Wang. 2023.
\newblock \href {https://api.semanticscholar.org/CorpusID:258865395} {Grace: Discriminator-guided chain-of-thought reasoning}.
\newblock In \emph{Conference on Empirical Methods in Natural Language Processing}.

\bibitem[{Kocsis and Szepesv{\'a}ri(2006)}]{kocsis2006bandit}
Levente Kocsis and Csaba Szepesv{\'a}ri. 2006.
\newblock Bandit based monte-carlo planning.
\newblock In \emph{European conference on machine learning}, pages 282--293. Springer.

\bibitem[{Ladosz et~al.(2022)Ladosz, Weng, Kim, and Oh}]{ladosz2022exploration}
Pawel Ladosz, Lilian Weng, Minwoo Kim, and Hyondong Oh. 2022.
\newblock Exploration in deep reinforcement learning: A survey.
\newblock \emph{Information Fusion}, 85:1--22.

\bibitem[{Li et~al.(2023)Li, Wang, Zhu, Zhang, Hou, Lian, and Xie}]{li2023emotionprompt}
Cheng Li, Jindong Wang, Kaijie Zhu, Yixuan Zhang, Wenxin Hou, Jianxun Lian, and Xing Xie. 2023.
\newblock Emotionprompt: Leveraging psychology for large language models enhancement via emotional stimulus.
\newblock \emph{arXiv preprint arXiv:2307.11760}.

\bibitem[{Liu et~al.(2023{\natexlab{a}})Liu, Lin, Hewitt, Paranjape, Bevilacqua, Petroni, and Liang}]{DBLP:journals/corr/abs-2307-03172}
Nelson~F. Liu, Kevin Lin, John Hewitt, Ashwin Paranjape, Michele Bevilacqua, Fabio Petroni, and Percy Liang. 2023{\natexlab{a}}.
\newblock \href {https://doi.org/10.48550/ARXIV.2307.03172} {Lost in the middle: How language models use long contexts}.
\newblock \emph{CoRR}, abs/2307.03172.

\bibitem[{Liu et~al.(2023{\natexlab{b}})Liu, Yang, Jia, Zhang, Zhou, Dai, Yang, and Vosoughi}]{liu2023training}
Ruibo Liu, Ruixin Yang, Chenyan Jia, Ge~Zhang, Denny Zhou, Andrew~M Dai, Diyi Yang, and Soroush Vosoughi. 2023{\natexlab{b}}.
\newblock Training socially aligned language models in simulated human society.
\newblock \emph{arXiv preprint arXiv:2305.16960}.

\bibitem[{Madaan et~al.(2023)Madaan, Tandon, Gupta, Hallinan, Gao, Wiegreffe, Alon, Dziri, Prabhumoye, Yang, Welleck, Majumder, Gupta, Yazdanbakhsh, and Clark}]{DBLP:journals/corr/abs-2303-17651}
Aman Madaan, Niket Tandon, Prakhar Gupta, Skyler Hallinan, Luyu Gao, Sarah Wiegreffe, Uri Alon, Nouha Dziri, Shrimai Prabhumoye, Yiming Yang, Sean Welleck, Bodhisattwa~Prasad Majumder, Shashank Gupta, Amir Yazdanbakhsh, and Peter Clark. 2023.
\newblock \href {https://doi.org/10.48550/ARXIV.2303.17651} {Self-refine: Iterative refinement with self-feedback}.
\newblock \emph{CoRR}, abs/2303.17651.

\bibitem[{Manakul et~al.(2023)Manakul, Liusie, and Gales}]{DBLP:journals/corr/abs-2303-08896}
Potsawee Manakul, Adian Liusie, and Mark J.~F. Gales. 2023.
\newblock \href {https://doi.org/10.48550/ARXIV.2303.08896} {Selfcheckgpt: Zero-resource black-box hallucination detection for generative large language models}.
\newblock \emph{CoRR}, abs/2303.08896.

\bibitem[{Marks and Tegmark(2023)}]{Marks2023TheGO}
Samuel Marks and Max Tegmark. 2023.
\newblock \href {https://api.semanticscholar.org/CorpusID:263831277} {The geometry of truth: Emergent linear structure in large language model representations of true/false datasets}.
\newblock \emph{ArXiv}, abs/2310.06824.

\bibitem[{Min et~al.(2022)Min, Lyu, Holtzman, Artetxe, Lewis, Hajishirzi, and Zettlemoyer}]{DBLP:conf/emnlp/MinLHALHZ22}
Sewon Min, Xinxi Lyu, Ari Holtzman, Mikel Artetxe, Mike Lewis, Hannaneh Hajishirzi, and Luke Zettlemoyer. 2022.
\newblock \href {https://doi.org/10.18653/V1/2022.EMNLP-MAIN.759} {Rethinking the role of demonstrations: What makes in-context learning work?}
\newblock In \emph{Proceedings of the 2022 Conference on Empirical Methods in Natural Language Processing, {EMNLP} 2022, Abu Dhabi, United Arab Emirates, December 7-11, 2022}, pages 11048--11064. Association for Computational Linguistics.

\bibitem[{Mu et~al.(2022)Mu, Zhong, Raileanu, Jiang, Goodman, Rockt{\"a}schel, and Grefenstette}]{mu2022improving}
Jesse Mu, Victor Zhong, Roberta Raileanu, Minqi Jiang, Noah Goodman, Tim Rockt{\"a}schel, and Edward Grefenstette. 2022.
\newblock Improving intrinsic exploration with language abstractions.
\newblock \emph{Advances in Neural Information Processing Systems}, 35:33947--33960.

\bibitem[{Oudeyer and Kaplan(2007)}]{oudeyer2007intrinsic}
Pierre-Yves Oudeyer and Frederic Kaplan. 2007.
\newblock What is intrinsic motivation? a typology of computational approaches.
\newblock \emph{Frontiers in neurorobotics}, 1:6.

\bibitem[{Pan et~al.(2023)Pan, Saxon, Xu, Nathani, Wang, and Wang}]{Pan2023AutomaticallyCL}
Liangming Pan, Michael~Stephen Saxon, Wenda Xu, Deepak Nathani, Xinyi Wang, and William~Yang Wang. 2023.
\newblock \href {https://api.semanticscholar.org/CorpusID:260682695} {Automatically correcting large language models: Surveying the landscape of diverse self-correction strategies}.
\newblock \emph{ArXiv}, abs/2308.03188.

\bibitem[{Parthasarathy et~al.(2023)Parthasarathy, Kontes, Plinge, and Mutschler}]{DBLP:journals/corr/abs-2305-16209}
Dinesh Parthasarathy, Georgios~D. Kontes, Axel Plinge, and Christopher Mutschler. 2023.
\newblock \href {https://doi.org/10.48550/ARXIV.2305.16209} {{C-MCTS:} safe planning with monte carlo tree search}.
\newblock \emph{CoRR}, abs/2305.16209.

\bibitem[{Paul et~al.(2023)Paul, Ismayilzada, Peyrard, Borges, Bosselut, West, and Faltings}]{Paul2023REFINERRF}
Debjit Paul, Mete Ismayilzada, Maxime Peyrard, Beatriz Borges, Antoine Bosselut, Robert West, and Boi Faltings. 2023.
\newblock \href {https://api.semanticscholar.org/CorpusID:257921623} {Refiner: Reasoning feedback on intermediate representations}.
\newblock \emph{ArXiv}, abs/2304.01904.

\bibitem[{Peng et~al.(2023)Peng, Galley, He, Cheng, Xie, Hu, Huang, Liden, Yu, Chen, and Gao}]{DBLP:journals/corr/abs-2302-12813}
Baolin Peng, Michel Galley, Pengcheng He, Hao Cheng, Yujia Xie, Yu~Hu, Qiuyuan Huang, Lars Liden, Zhou Yu, Weizhu Chen, and Jianfeng Gao. 2023.
\newblock \href {https://doi.org/10.48550/ARXIV.2302.12813} {Check your facts and try again: Improving large language models with external knowledge and automated feedback}.
\newblock \emph{CoRR}, abs/2302.12813.

\bibitem[{Raffel et~al.(2020)Raffel, Shazeer, Roberts, Lee, Narang, Matena, Zhou, Li, and Liu}]{JMLR:v21:20-074}
Colin Raffel, Noam Shazeer, Adam Roberts, Katherine Lee, Sharan Narang, Michael Matena, Yanqi Zhou, Wei Li, and Peter~J. Liu. 2020.
\newblock \href {http://jmlr.org/papers/v21/20-074.html} {Exploring the limits of transfer learning with a unified text-to-text transformer}.
\newblock \emph{Journal of Machine Learning Research}, 21(140):1--67.

\bibitem[{Shinn et~al.(2023)Shinn, Cassano, Labash, Gopinath, Narasimhan, and Yao}]{Shinn2023ReflexionLA}
Noah Shinn, Federico Cassano, Beck Labash, Ashwin Gopinath, Karthik Narasimhan, and Shunyu Yao. 2023.
\newblock \href {https://api.semanticscholar.org/CorpusID:258833055} {Reflexion: Language agents with verbal reinforcement learning}.

\bibitem[{Swiechowski et~al.(2023)Swiechowski, Godlewski, Sawicki, and Mandziuk}]{DBLP:journals/air/SwiechowskiGSM23}
Maciej Swiechowski, Konrad Godlewski, Bartosz Sawicki, and Jacek Mandziuk. 2023.
\newblock \href {https://doi.org/10.1007/S10462-022-10228-Y} {Monte carlo tree search: a review of recent modifications and applications}.
\newblock \emph{Artif. Intell. Rev.}, 56(3):2497--2562.

\bibitem[{Thorne et~al.(2018)Thorne, Vlachos, Christodoulopoulos, and Mittal}]{Thorne18Fever}
James Thorne, Andreas Vlachos, Christos Christodoulopoulos, and Arpit Mittal. 2018.
\newblock {FEVER}: a large-scale dataset for fact extraction and {VERification}.
\newblock In \emph{NAACL-HLT}.

\bibitem[{Touvron et~al.(2023)Touvron, Martin, Stone, Albert, Almahairi, Babaei, Bashlykov, Batra, Bhargava, Bhosale, Bikel, Blecher, Ferrer, Chen, Cucurull, Esiobu, Fernandes, Fu, Fu, Fuller, Gao, Goswami, Goyal, Hartshorn, Hosseini, Hou, Inan, Kardas, Kerkez, Khabsa, Kloumann, Korenev, Koura, Lachaux, Lavril, Lee, Liskovich, Lu, Mao, Martinet, Mihaylov, Mishra, Molybog, Nie, Poulton, Reizenstein, Rungta, Saladi, Schelten, Silva, Smith, Subramanian, Tan, Tang, Taylor, Williams, Kuan, Xu, Yan, Zarov, Zhang, Fan, Kambadur, Narang, Rodriguez, Stojnic, Edunov, and Scialom}]{Touvron2023Llama2O}
Hugo Touvron, Louis Martin, Kevin~R. Stone, Peter Albert, Amjad Almahairi, Yasmine Babaei, Nikolay Bashlykov, Soumya Batra, Prajjwal Bhargava, Shruti Bhosale, Daniel~M. Bikel, Lukas Blecher, Cristian~Cant{\'o}n Ferrer, Moya Chen, Guillem Cucurull, David Esiobu, Jude Fernandes, Jeremy Fu, Wenyin Fu, Brian Fuller, Cynthia Gao, Vedanuj Goswami, Naman Goyal, Anthony~S. Hartshorn, Saghar Hosseini, Rui Hou, Hakan Inan, Marcin Kardas, Viktor Kerkez, Madian Khabsa, Isabel~M. Kloumann, A.~V. Korenev, Punit~Singh Koura, Marie-Anne Lachaux, Thibaut Lavril, Jenya Lee, Diana Liskovich, Yinghai Lu, Yuning Mao, Xavier Martinet, Todor Mihaylov, Pushkar Mishra, Igor Molybog, Yixin Nie, Andrew Poulton, Jeremy Reizenstein, Rashi Rungta, Kalyan Saladi, Alan Schelten, Ruan Silva, Eric~Michael Smith, R.~Subramanian, Xia Tan, Binh Tang, Ross Taylor, Adina Williams, Jian~Xiang Kuan, Puxin Xu, Zhengxu Yan, Iliyan Zarov, Yuchen Zhang, Angela Fan, Melanie Kambadur, Sharan Narang, Aurelien Rodriguez, Robert Stojnic, Sergey Edunov, and
  Thomas Scialom. 2023.
\newblock \href {https://api.semanticscholar.org/CorpusID:259950998} {Llama 2: Open foundation and fine-tuned chat models}.
\newblock \emph{ArXiv}, abs/2307.09288.

\bibitem[{Wang et~al.(2023)Wang, Wei, Schuurmans, Le, Chi, Narang, Chowdhery, and Zhou}]{DBLP:conf/iclr/0002WSLCNCZ23}
Xuezhi Wang, Jason Wei, Dale Schuurmans, Quoc~V. Le, Ed~H. Chi, Sharan Narang, Aakanksha Chowdhery, and Denny Zhou. 2023.
\newblock \href {https://openreview.net/pdf?id=1PL1NIMMrw} {Self-consistency improves chain of thought reasoning in language models}.
\newblock In \emph{The Eleventh International Conference on Learning Representations, {ICLR} 2023, Kigali, Rwanda, May 1-5, 2023}. OpenReview.net.

\bibitem[{Wei et~al.(2022)Wei, Wang, Schuurmans, Bosma, brian ichter, Xia, Chi, Le, and Zhou}]{wei2022chain}
Jason Wei, Xuezhi Wang, Dale Schuurmans, Maarten Bosma, brian ichter, Fei Xia, Ed~H. Chi, Quoc~V Le, and Denny Zhou. 2022.
\newblock \href {https://openreview.net/forum?id=_VjQlMeSB_J} {Chain of thought prompting elicits reasoning in large language models}.
\newblock In \emph{Advances in Neural Information Processing Systems}.

\bibitem[{Xie et~al.(2023{\natexlab{a}})Xie, Zhang, Chen, Lou, and Su}]{Xie2023AdaptiveCO}
Jian Xie, Kai Zhang, Jiangjie Chen, Renze Lou, and Yu~Su. 2023{\natexlab{a}}.
\newblock \href {https://api.semanticscholar.org/CorpusID:263610324} {Adaptive chameleon or stubborn sloth: Revealing the behavior of large language models in knowledge conflicts}.

\bibitem[{Xie et~al.(2023{\natexlab{b}})Xie, Kawaguchi, Zhao, Zhao, Kan, He, and Xie}]{Xie2023SelfEvaluationGB}
Yuxi Xie, Kenji Kawaguchi, Yiran Zhao, Xu~Zhao, MingSung Kan, Junxian He, and Qizhe Xie. 2023{\natexlab{b}}.
\newblock \href {https://api.semanticscholar.org/CorpusID:258426922} {Self-evaluation guided beam search for reasoning}.

\bibitem[{Xu et~al.(2023)Xu, Xu, Wang, Liu, Zhu, and McAuley}]{DBLP:journals/corr/abs-2305-08848}
Canwen Xu, Yichong Xu, Shuohang Wang, Yang Liu, Chenguang Zhu, and Julian~J. McAuley. 2023.
\newblock \href {https://doi.org/10.48550/ARXIV.2305.08848} {Small models are valuable plug-ins for large language models}.
\newblock \emph{CoRR}, abs/2305.08848.

\bibitem[{Yao et~al.(2023{\natexlab{a}})Yao, Yu, Zhao, Shafran, Griffiths, Cao, and Narasimhan}]{Yao2023TreeOT}
Shunyu Yao, Dian Yu, Jeffrey Zhao, Izhak Shafran, Thomas~L. Griffiths, Yuan Cao, and Karthik Narasimhan. 2023{\natexlab{a}}.
\newblock \href {https://api.semanticscholar.org/CorpusID:258762525} {Tree of thoughts: Deliberate problem solving with large language models}.
\newblock \emph{ArXiv}, abs/2305.10601.

\bibitem[{Yao et~al.(2023{\natexlab{b}})Yao, Zhao, Yu, Du, Shafran, Narasimhan, and Cao}]{DBLP:conf/iclr/YaoZYDSN023}
Shunyu Yao, Jeffrey Zhao, Dian Yu, Nan Du, Izhak Shafran, Karthik~R. Narasimhan, and Yuan Cao. 2023{\natexlab{b}}.
\newblock \href {https://openreview.net/pdf?id=WE\_vluYUL-X} {React: Synergizing reasoning and acting in language models}.
\newblock In \emph{The Eleventh International Conference on Learning Representations, {ICLR} 2023, Kigali, Rwanda, May 1-5, 2023}. OpenReview.net.

\bibitem[{Zelikman et~al.(2022)Zelikman, Wu, Mu, and Goodman}]{zelikman2022star}
Eric Zelikman, Yuhuai Wu, Jesse Mu, and Noah Goodman. 2022.
\newblock \href {https://openreview.net/forum?id=_3ELRdg2sgI} {{ST}ar: Bootstrapping reasoning with reasoning}.
\newblock In \emph{Advances in Neural Information Processing Systems}.

\bibitem[{Zhang et~al.(2023)Zhang, Luo, Chuang, Fang, Gaitskell, Hartvigsen, Wu, Fox, Meng, and Glass}]{Zhang2023InterpretableUL}
Tianhua Zhang, Hongyin Luo, Yung-Sung Chuang, Wei Fang, Luc Gaitskell, Thomas Hartvigsen, Xixin Wu, Danny Fox, Helen~M. Meng, and James~R. Glass. 2023.
\newblock \href {https://api.semanticscholar.org/CorpusID:258041307} {Interpretable unified language checking}.
\newblock \emph{ArXiv}, abs/2304.03728.

\bibitem[{Zheng et~al.(2023)Zheng, Chiang, Sheng, Zhuang, Wu, Zhuang, Lin, Li, Li, Xing, Zhang, Gonzalez, and Stoica}]{Zheng2023JudgingLW}
Lianmin Zheng, Wei-Lin Chiang, Ying Sheng, Siyuan Zhuang, Zhanghao Wu, Yonghao Zhuang, Zi~Lin, Zhuohan Li, Dacheng Li, Eric~P. Xing, Haotong Zhang, Joseph Gonzalez, and Ion Stoica. 2023.
\newblock \href {https://api.semanticscholar.org/CorpusID:259129398} {Judging llm-as-a-judge with mt-bench and chatbot arena}.
\newblock \emph{ArXiv}, abs/2306.05685.

\bibitem[{Zhou et~al.(2023)Zhou, Yan, Shlapentokh{-}Rothman, Wang, and Wang}]{DBLP:journals/corr/abs-2310-04406}
Andy Zhou, Kai Yan, Michal Shlapentokh{-}Rothman, Haohan Wang, and Yu{-}Xiong Wang. 2023.
\newblock \href {https://doi.org/10.48550/ARXIV.2310.04406} {Language agent tree search unifies reasoning acting and planning in language models}.
\newblock \emph{CoRR}, abs/2310.04406.

\bibitem[{Zhu et~al.(2023)Zhu, Wang, Zhang, Zhang, Huang, Gan, Zhang, and Yang}]{zhu-etal-2023-solving}
Xinyu Zhu, Junjie Wang, Lin Zhang, Yuxiang Zhang, Yongfeng Huang, Ruyi Gan, Jiaxing Zhang, and Yujiu Yang. 2023.
\newblock \href {https://doi.org/10.18653/v1/2023.acl-long.245} {Solving math word problems via cooperative reasoning induced language models}.
\newblock In \emph{Proceedings of the 61st Annual Meeting of the Association for Computational Linguistics (Volume 1: Long Papers)}, pages 4471--4485, Toronto, Canada. Association for Computational Linguistics.

\end{thebibliography}

\clearpage
\appendix
\setcounter{table}{0}
\renewcommand{\thetable}{A\arabic{table}}

\setcounter{figure}{0}
\renewcommand{\thefigure}{A\arabic{figure}}

\section{More Experimental Details for Initial Study}
\subsection{Experiment for Figure~\ref{fig:fail_self_asses}}
\label{app:Autostop}
The prompt used in \textsl{Autostop} is \texttt{``You were either successful or unsuccessful in your previous trial. Stick to your previous answer if it is correct, otherwise consider a new answer''}. The prompt used for \textsl{NeverStop} is \texttt{``You failed in your previous trial and reconsider a new answer''}.  The motivation behind \textsl{Autostop} is that we totally rely on the LLM's internal knowledge to check the correctness of its own outputs. However, LLM fails in this setting as the performance is even worse than initial stage. For \textsl{NeverStop}, we hope to identify that some correctly answered samples will be kept unchanged even the negative feedback provided. However, we didn't find a pattern between the changed and unchanged predicted samples. 

\subsection{Implementation for Knowledge Grounding and Results}
\label{app:kg_ex}
\paragraph{Dataset} We evaluate LLMs' knowledge grounding ability on knowledge-rich multiple-choice dataset, MMLU. It consists of four domains: STEM, Social, Humanity and Other, totaling 56 subjects. All methods are evaluated on 50 randomly selected samples for each subject (excluding those in the Other domain), and the remaining samples are used as the training set where applicable. 
\paragraph{Models and Baselines} In addition to Llama2-13B, Llama2-70B, and GPT-3.5 for prompting, we also leverage unified language checking, \textsl{UniLangCheck}~\citep{Zhang2023InterpretableUL}, for statement assessment. \textsl{UniLangCheck} aims to check if language input is factual and fair via prompting LLMs to generate groundings for fact-checking. Therefore, we firstly prompt LLMs to generate a fact about the key element in the question before proceeding to the final assessment. We repeatedly prompt the LLMs for 5 times and use the majority-voted answer  as the result for \textsl{Self-Consistency}~\citep{DBLP:conf/iclr/0002WSLCNCZ23}. \textsl{TRUE}~\citep{honovich-etal-2022-true-evaluating} is the T5-11B~\cite{JMLR:v21:20-074} model fine-tuned on a collection of
natural language inference (NLI) datasets to check factual correctness, and has been used by previous works within similar
contexts~\citep{gao-etal-2023-rarr,gao2023enabling}. We further fine-tune its classifier head on our training set, which is annotated as factually correct or not, before evaluation. Both Contrastive Consistent Search (\textsl{ContrastSearch})~\citep{DBLP:conf/iclr/BurnsYKS23} and 
\textsl{ActivationRegress}~\citep{Marks2023TheGO} train classifiers whose inputs are activations extracted from Llama2-13B 12-layer encodings~\footnote{The original dimensions of Llama2-30B is 5024. We apply PCA to reduce this dimensionality to obtain 50-dimensional activations as classifier input.}.
\textsl{ActivationRegress} trains a logistic classifier on the activations with factual labels as supervision. \textsl{ContrastSearch}, instead, operates without factual labels. For a statement $s_i$, we firstly construct a data-pair $x^{+}$ and $x^{-}$ by annotating \textit{True} and \textit{False} to this statement ,regardless of its factual correctness. Then, we derive the probabilities by mapping $x$ to a number between 0 and 1, i.e., $p^{+}=p_{\theta}(\phi(x_{i}^{+}))$ and $p^{-}=p_{\theta}(\phi(x_{i}^{-}))$. The mapping function $p_{\theta}$ is updated such that the
probabilities are both confident ($p_{i}^{+}\approx 1-p_{i}^{-}$) and consistent ($p_{i}^{+}\not\approx p_{i}^{-}$). 

\begin{figure*}
\centering
\begin{subfigure}[t]{0.95\textwidth}
   \includegraphics[trim={10 400 130 10}, clip,width=0.95\textwidth]{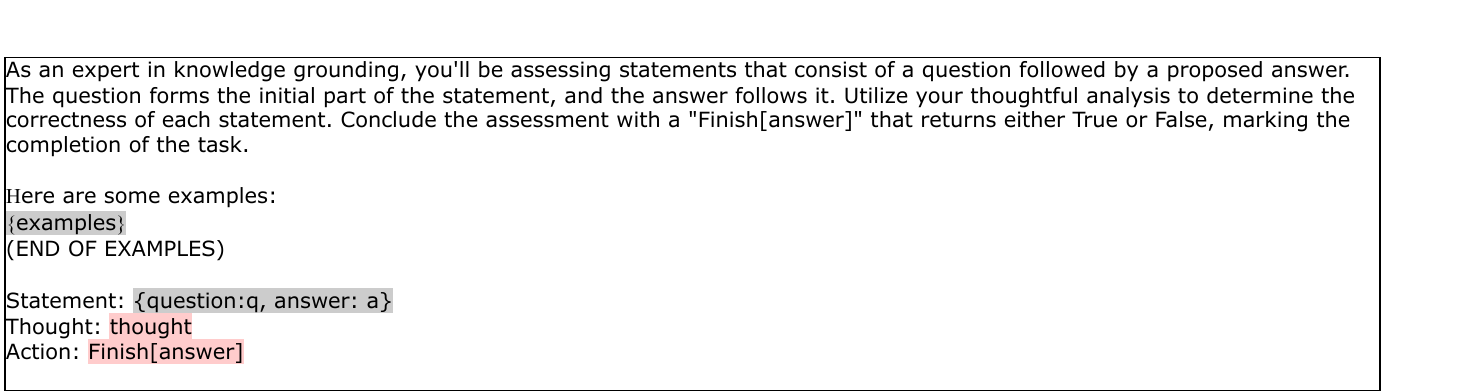}
   \caption{Basic prompt for knowledge grounding. Text in gray is extracted from datasets, in red shadow is generated by LLMs.}
   \label{fig:basic_prompt_kg} 
\end{subfigure}
\vfill
\begin{subfigure}[t]{0.95\textwidth}
   \includegraphics[trim={10 130 130 260}, clip,width=0.95\textwidth]{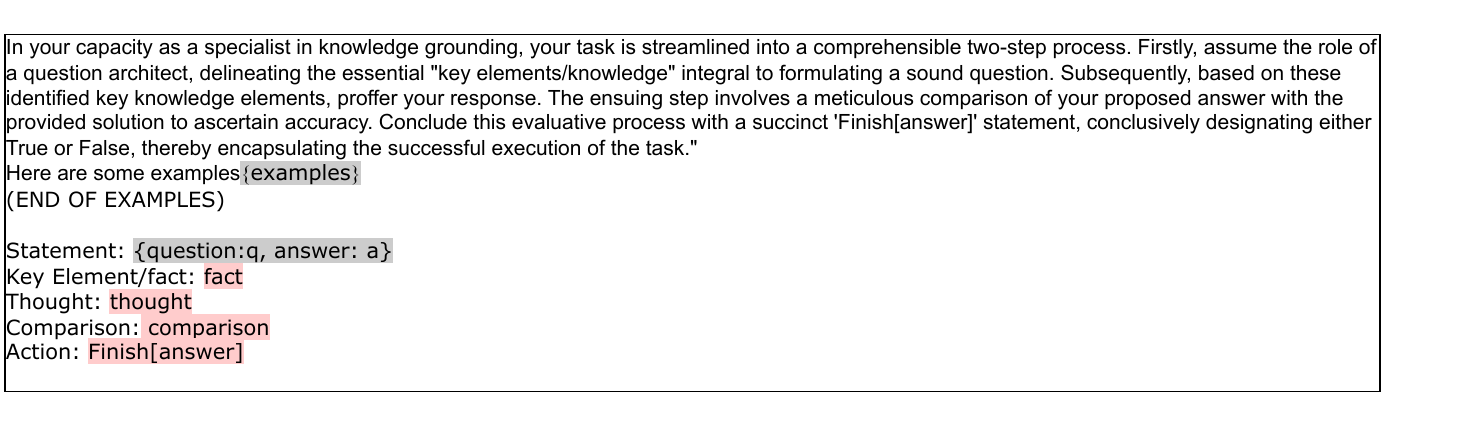}
   \caption{Fact-extract prompt applied to \textsl{UniLangCheck} for knowledge grounding. Text in gray shadow is extracted from datasets, in red shadow is generated by LLMs. Comparing to the basic prompt, it includes additional fact generation.}
   \label{fig:unilc_prompt_kg} 
\end{subfigure}
\caption{Prompts for knowledge grounding.}
\end{figure*}

\paragraph{Prompt Settings} The basic prompt for knowledge grounding is shown in Figure~\ref{fig:basic_prompt_kg}. This is used for Llama2, GPT-3.5 and Self-Consistency. The advanced prompt inspired by \textsl{UniLangCheck} is illustrated in Figure~\ref{fig:unilc_prompt_kg}. For each subject, we randomly select 50 samples and extract their question and choice to build a statement for knowledge checking. The correctness of this statement is deemed \textsl{True} if the selected choice is exactly the correct one, otherwise it is labeled \textsl{False}.

\paragraph{Correlation between Self-consistency Confidence and Accuracy}
For the self-consistency(5) baseline, we calculate the $\mathcal{R}^{2}$ for confidence (the frequency of the current answer among all generated answers, totaling 5) and the accuracy. The results are shown in Table~\ref{tab:acc_consistency}. We observe a high correlation between the two variables, which inspires our design of multiple-consistency for answer assessment. 
\begin{table}[h]
    \centering
    \resizebox{0.48\textwidth}{!}{%
    \begin{tabular}{crrrr}
    \toprule[1pt]
         & STEM & Social & Humanity & Others \\
         \cline{2-5}
   GPT-3.5&   0.80 & 0.89 & 0.84& 0.88     \\
   Llama& 0.86&0.85& 0.91& 0.86\\
   Vicuna& 0.92&0.90 & 0.74&0.92\\
    \bottomrule[1pt]
    \end{tabular}
    }
    \caption{The correlation between accuracy and self-consistency confidence over three domains in the MMLU datasets.}
    \label{tab:acc_consistency}
\end{table}



\subsection{Implementation for Direction Generation}
\begin{figure*}[t]
    \includegraphics[trim={2 225 2 10}, clip, width=0.98\textwidth]{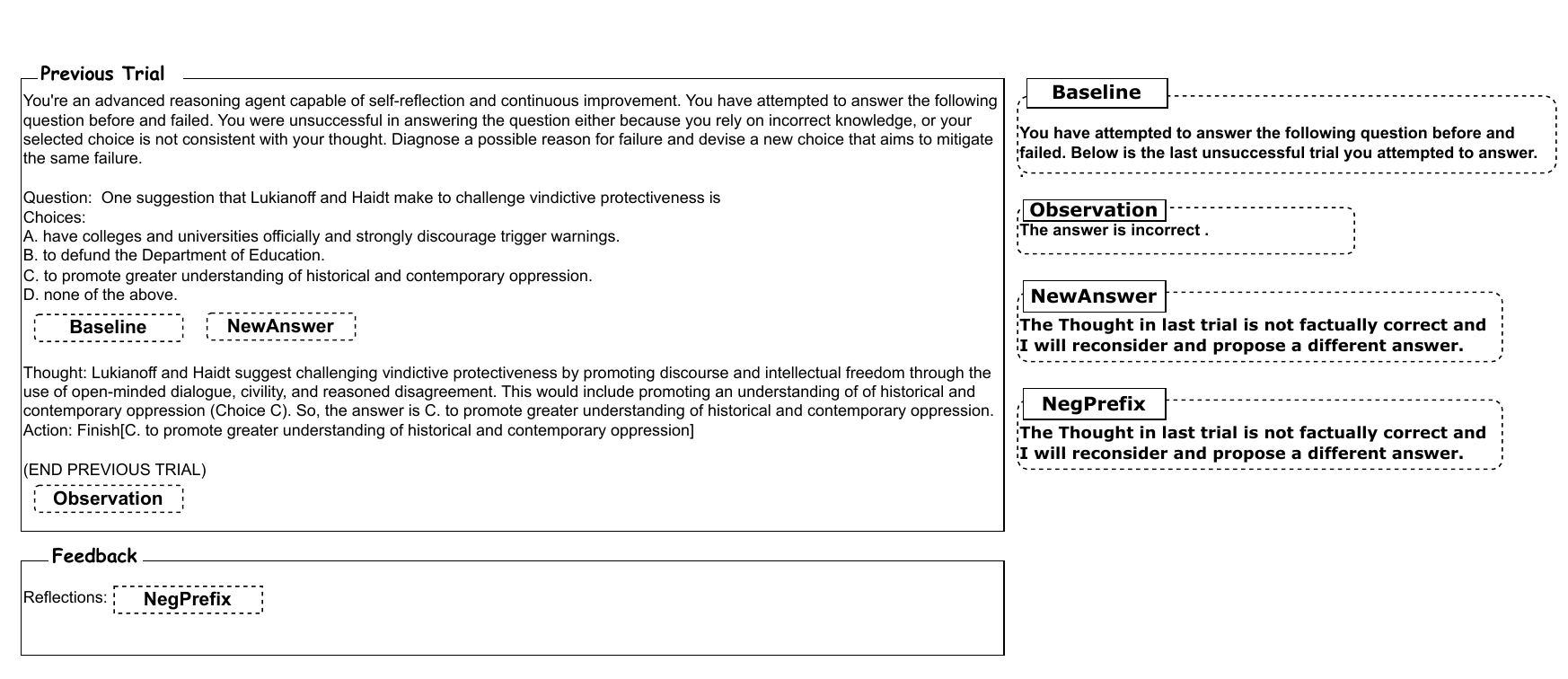}
    \caption{Given the question and the LLM's \textsl{previous trial}, it is asked to generated \textsl{feedback} under different prompts to facilitate reflection and potentially update its previous response. The four candidate instructions, Baseline, Observation, NewAnswer and NegPrefix, are enclosed in dashed frames, and they will be positioned differently to exert their respective effects. 
    }
    \label{fig:alignment_prompts}
\end{figure*}

Based on the observation that existing feedback has limited effects to guide LLMs to update their current incorrect response, we propose several simple strategies to enhance the effectiveness of generated feedback in the self-improvement process. These strategies are mainly inspired by the following two observations: (1) LLMs are more susceptible to context influence at the beginning or near the end~\cite{DBLP:journals/corr/abs-2307-03172} (2) ICL is highly sensitive to the stylish and emotional words in  demonstrations~\cite{DBLP:conf/emnlp/MinLHALHZ22,li2023emotionprompt}.
We summarize the different strategies in the diagram shown in Figure~\ref{fig:alignment_prompts}. 

We show relative percentage of changed samples those incorrectly predicted ones before and after applying the \textsl{NegReflect} in Table~\ref{tab:ans_changes_after}. The percentages have been greatly improved with the instruction which has been inserted closer to the end of prompt. To verify whether this change could lead to task performance, we display the detailed performances over three LLMs after applying different instructions in Table~\ref{tab:follow_inst}. It is clear that \textsl{NegPrefix} demonstrates the most significant improvements across all the datasets and models. In contrast, \textsl{NewAnswer} has the same sentences \textsl{NegPrefix} as but its position is far away from the generating point for LLMs. \textit{This can be explained that position of instruction is important in ICL.} And the performance of \textsl{NewAnswer} is slightly better than baseline, it can be partly explained that the \textsl{NewAnswer} explicitly show the negative attitude towards and guide the model to generate a different answer. Among the three models, the average promotion on GPT3.5 is the most negligible. \textit{This can be explained that larger model are more confident with its internal knowledge and less vulnerable to given noisy text.} 
\begin{table}[ht]
    \centering
    \resizebox{0.35\textwidth}{!}{%
    \begin{tabular}{lrc}
    \toprule[1pt]
    \textbf{Model}     & \textbf{Prompts} &  \textbf{Change} \\
\midrule
    \multirow{2}{*}{GPT35} & Oracle &0.56  \\
    & NegReflect & 0.72\\
    \midrule
        \multirow{2}{*}{Llama} & Oracle &0.54  \\
    & NegReflect & 0.72\\
    \midrule
        \multirow{2}{*}{Vicuna} & Oracle  & 0.64 \\
    & NegReflect & 0.74\\
    \bottomrule[1pt]
    \end{tabular}
    }
    \caption{The relative percentage of changed samples among those incorrectly predicted ones. We use the average results for different domains in MMLU. }
    \label{tab:ans_changes_after}
\end{table}

\begin{table}[ht]
    \centering
    \resizebox{0.45\textwidth}{!}{%
    \begin{tabular}{lrrrr}
    \toprule[1pt]
      Model   & Stem&Social&Humanity&Other  \\
      \hline
     GPT35    & 0.80
     &0.82&0.78&0.73\\
     +Observation&0.76&0.82&0.75&0.70 \\
     +NewAnswer&0.80&0.84&0.80&0.75 \\
     +NegReflect&0.84&0.86&0.84&0.76\\
     \hline 
          Llama    &0.64&0.63&0.60&0.64 \\
     +Observation&0.63&0.62&0.61&0.62\\
     +NewAnswer&0.64&0.67&0.65&0.64\\
     +NegReflect&0.70&0.72&0.76&0.69\\
     \hline 
    Vicuna    &0.62&0.68&0.59&0.69\\
     +Observation&0.64&0.52&0.45&0.67\\
     +NewAnswer&0.66&0.58&0.47&0.65\\
     +NegReflect&0.69&0.63&0.52&0.72\\
     \bottomrule[1pt]
    \end{tabular}
    }
    \caption{Self-improvments results with different prompt constraints for answer correction. By comparing with the ground truth, this evaluation is to show the capability of LLMs in obeying the instructions of changing their incorrect predictions.}
    \label{tab:follow_inst}
\end{table}


\section{Mirror algorithm}
\label{app:mcts}
We introduce the pipeline of the proposed Mirror in Algorithm~\ref{alg:cap} involves iteratively conducting a \textsc{UCT-Search} until predefined iteration constraint is reached, and the best action $a(\textsc{BestChild}(v_0,0))$ leading to the best child of the root node $v_0$ returns. Node in the tree is $v$ and its associated state is $s(v)$, representing the response generated by Reasoner. The action is $a(v)$, reward is $\mathcal{R}$ and $N(\cdot)$ is the times of the node having been visited. $r(v)$ is the reward for the terminate state at each iteration. 

The overall process consists of three steps:
(1) \textsc{SearchPolicy} to obtain the terminal node $v_l$. through  which expands the tree until fully expanded. Specially, we randomly add one or more nodes to the root node according to the possible actions. In our case, we generate multiple responses to the given question and previous attempts/response. When the current node is fully expanded, we apply the UTC algorithm to select the best child node. 
(2) \textsc{Simulation} the reward $r$ for $v_l$ through \textsc{SimulationPolicy}. This phrase is to simulate the future rewards of the current node through multiple interactions. For simplicity, we follow the similar process as expansion and return the reward $r$ for selected action-state pair.
(3) \textsc{BackPropagate} the simulation results to the selected nodes to accelerate \textsc{SearchPolicy} in next iteration.
Algorithm is described in Algorithm~\ref{alg:cap}.

\begin{algorithm}[h]
\caption{\texttt{Mirror}-UCT}\label{alg:cap}
\begin{minipage}{0.89\linewidth} 
\begin{algorithmic}
\scriptsize
\Require state transition function $f : S \times A \rightarrow S$, weight $C_p$, Reward $\mathcal{R}$, Stop Criteria $g\rightarrow\{0,1\}$
\State \textbf{function} \textsc{uct-search}$(s_0)$ 
    \State \hspace{\algorithmicindent}  create root node $v_0$ with state $s_0$
    \State \hspace{\algorithmicindent} \textbf{while} {within computational iteration} \textbf{do}
        \State \hspace{\algorithmicindent} \hspace{\algorithmicindent} $v_l \leftarrow $ \textsc{SearchPolicy}$(v_0)$ 
        \State \hspace{\algorithmicindent} \hspace{\algorithmicindent} $r \leftarrow \textsc{SimulationPolicy}(s_{v_l})$
        \State \hspace{\algorithmicindent} \hspace{\algorithmicindent} $\textsc{BackPropagate}(v_l,r)$
        \State \hspace{\algorithmicindent} \textbf{return} $a(\textsc{BestChild}(v_0,0))$
        
\State

\State \textbf{function} $\textsc{SearchPolicy}(v)$
\State \hspace{\algorithmicindent} \textbf{while} $g(v) == 0$ \textbf{do}
\State \hspace{\algorithmicindent} \hspace{\algorithmicindent} \textbf{if} $v$ not fully expanded \textbf{then}
\State \hspace{\algorithmicindent} \hspace{\algorithmicindent} \hspace{\algorithmicindent} \textbf{return} 
\textsc{Expand}$(v)$
\State \hspace{\algorithmicindent} \hspace{\algorithmicindent} \textbf{else} 
\State \hspace{\algorithmicindent} \hspace{\algorithmicindent} \hspace{\algorithmicindent} $v \leftarrow \textsc{BestChild}(v,C_p)$
\State \hspace{\algorithmicindent} \hspace{\algorithmicindent} \textbf{return} $v$

\State

\State \textbf{function} \textsc{Expand}$(v)$
\State \hspace{\algorithmicindent} choose $a\in$ untried actions from $A(s(v))$
\State \hspace{\algorithmicindent} add a new child $v'$ to $v$
\State  \hspace{\algorithmicindent}  \hspace{\algorithmicindent} with $s(v)' = f(s(v),a)$ and $a(v') = a$
\State  \hspace{\algorithmicindent} \textbf{return} $v'$

\State 
\State \textbf{function} \textsc{BestChild}$(v,C_p)$
\State \hspace{\algorithmicindent} \textbf{return} $\underset{v'\in \text{children of}\ v}{\text{argmax}}$ $\frac{{\mathcal{R}}_{v'}}{N(v')} + 2C_p\sqrt{\frac{2\ \text{ln}\ N(v)}{N(v')}}$

\State
\State

\State \textbf{function} \textsc{SimulationPolicy}$(s)$
\State \hspace{\algorithmicindent} \textbf{While} $s$ is non-terminal \textbf{do}
\State \hspace{\algorithmicindent} \hspace{\algorithmicindent} $a = \underset{a\in A}{\text{argmax}}(\mathcal{R}(a,s))$ 
\State \hspace{\algorithmicindent} \hspace{\algorithmicindent} $s \leftarrow f(s,a)$
\State \hspace{\algorithmicindent} \textbf{return} reward for $s$
\State 

\State \textbf{function} \textsc{BackPropagate}$(v,r)$
\State \hspace{\algorithmicindent} \textbf{while} $v$ is not null \textbf{do}
\State \hspace{\algorithmicindent} \hspace{\algorithmicindent} $N(v) \leftarrow N(v)+1$
\State \hspace{\algorithmicindent} \hspace{\algorithmicindent} $\mathcal{R}(v) \leftarrow \mathcal{R}(v) + r(v)$
\State \hspace{\algorithmicindent} \hspace{\algorithmicindent} $v \leftarrow \text{parent of}\ v$

\end{algorithmic}
\end{minipage}
\end{algorithm}

\newpage
\section{Experiments for Mirror}
\label{app:mcts_ex}
We will introduce the implementation details and provide complementary results experimented on \texttt{Mirror} in this section.
\subsection{Implementation Details}
\label{app:inplement_details}
\paragraph{Hyper-parameter settings.} In order to encourage diverse direction generation, we set the generation temperature as 0.8 for all the models, and we set $\texttt{do}_{\_}\texttt{Sample}=\texttt{True}$ for llama and vicuna to avoid greedy search. For the threshold $T_0$ in self-assessment to deriving the final answer, we set 0.8 for GPT35, and 0.5 for llama and Vicuna according to the results on limited validation data. These results reveal that larger language models are more consistent in their multiple outputs, which is more difficult for the smaller models. Hence, we adopt a relatively lower threshold for smaller models. This observation can be partially explained by the tendency of larger LMs to rely on their parametric memory~\citep{Xie2023AdaptiveCO}.

\paragraph{Prompt Settings.} We provide 5 demonstrations along with instruction when prompting LLMs. We show the prompts/instructions provided to LLMs in direction generation and response generation process. \textbf{(a)} $p_0$ in direction generation in $\mathcal{\pi}(a_{t}|s_{t},p_0,\mathcal{R})$. The guidance in the upper is for initial response, the bottom one is for reflection in the subsequent iterations. \textbf{(b)} Prompt for response generation given previous response and direction. $\mathcal{P}(s_{t}|s_{t-1},a_{t-1};q)$.

\begin{tcolorbox}[enhanced,attach boxed title to top center={yshift=-3mm,yshifttext=-1mm}, colback=white,colframe=black,
  colbacktitle=black!75!yellow!,title=Prompt for Direction Generation (MMLU),boxed title style={size=small}]
\small As a tutor, your focus is on guiding the student to navigate multiple-choice question-answering problems strategically. Encourage them to dissect the question, identifying key elements and nuances within each choice. Emphasize the importance of understanding subtle differences that could distinguish correct from incorrect options.
\tcblower
\small As a tutor, your are supposed to meticulously evaluate the student's approach to multiple-choice problems.
Question, Choices and the student's previous thought and answer are given, check if the facts mentioned in the thought is correct and if there might be a more appropriate option than the one chosen. If the student's reasoning thought is accurate and the proposed answer is the most appropriate, encourage them to adhere to their initial trial. Otherwise, guide the student to revisit specific details, explore alternative choice.
\end{tcolorbox}

\begin{tcolorbox}[enhanced,attach boxed title to top center={yshift=-3mm,yshifttext=-1mm}, colback=white,colframe=black,
  colbacktitle=black!75!yellow!,title=Prompt for Direction Generation (FEVER),boxed title style={size=small}]
\small As a tutor, your focus is on guiding the student to navigate fact-checking problems strategically. Encourage them to dissect the claim, identifying key elements and associate facts. Emphasize the correct relation between important elements that could distinguish SUPPORTS from REFUTES options. Also, lacking of enough information will lead to NOT ENOUGH INFO.
\tcblower
\small As a tutor, your are supposed to meticulously evaluate the student's approach to fact verification task.
Claim and the student's previous thought and answer are given, check if the relations mentioned in the Thought is correct and if there might be a more appropriate answer. If the student's reasoning thought is accurate and the proposed answer is the most appropriate, encourage them to adhere to their initial trial. Otherwise, guide the student to revisit specific details, explore alternative answer.
\end{tcolorbox}

\begin{tcolorbox}[enhanced,attach boxed title to top center={yshift=-3mm,yshifttext=-1mm}, colback=white,colframe=black,
  colbacktitle=black!75!yellow!,title=Prompt for Response Generation (MMLU),boxed title style={size=small}]
\small You are an expert in multiple-choice question answering. Each problem will provide you with a question and answer choices. Read the question and all the choices carefully, along with the provided advice, and solve the problem by having a thought. Thought can reason about the current situation. Finish[answer] returns the answer and finishes the task.
\tcblower
\small You're an advanced reasoning agent capable of self-reflection and continuous improvement. Your objective is to tackle multiple-choice question answering problems. Each problem will provide you with a question, answer choices, your previous line of reasoning, and the detailed analyses from an experienced tutor. In a succinct review, assess the accuracy of your earlier answer based on your expertise and the advice, subsequently arrive at the definitive response.
\end{tcolorbox}

\begin{tcolorbox}[enhanced,attach boxed title to top center={yshift=-3mm,yshifttext=-1mm}, colback=white,colframe=black,
  colbacktitle=black!75!yellow!,title=Prompt for Response Generation (FEVER),boxed title style={size=small}]
\small You are a knowledgeable and accurate fact verifier. Please verify the correctness of the following claim based on your expertise and provided advice. Return SUPPORTS or REFUTES a Claim, or if there is NOT ENOUGH INFO.
\tcblower
\small You're an advanced reasoning agent capable of self-reflection in fact verification task. Claim and the your previous response and answer are given, along with the advice. In a succinct review, assess the accuracy of your earlier answer based on your expertise and the advice, subsequently arrive at the definitive response.
\end{tcolorbox}

\paragraph{Computational budget.}
The total running costs for using GPT-3.5 in our experiments are approximately \$500. 
In addition, the running time for Llama2 and Vicuna in our experiments totalled 320 hours, utilising one 40G A100 graphics cards.


\subsection{Additional Results}
We provide additional results as complementary to our main results.
\paragraph{Results on GSM8K dataset.} To further verify the effectiveness of Mirror in various reasoning tasks, we include the math problem, i.e., GSM8k~\citep{DBLP:journals/corr/abs-2110-14168}. The performance superiority is evident when comparing with the best-performing unsupervised baseline, i.e., self-consistency .
\begin{table}[h]
    \centering
    \resizebox{0.46\textwidth}{!}{%
    \begin{tabular}{cccc}
    \toprule
          COT & w. $\text{self-cons}^{(5)}$ & w. $\text{self-cons}^{(15)}$ & Mirror \\
          \midrule
         0.72 & 0.75 & 0.77 & 0.80 \\
        \bottomrule
    \end{tabular}
    }
    \caption{Based on GPT-3.5, performances of different reasoning methods.}
    \label{tab:gpt35_math}
\end{table}

\paragraph{Effects of question-oriented direction.}
To save computational resources, we randomly select 20 samples from each of the four domain datasets in MMLU and from FEVER.  We show an example of generated direction in the CoT settings. 
\begin{figure}[th]
    \centering
    \includegraphics[trim={0 320 510 0},clip,width=0.51\textwidth]{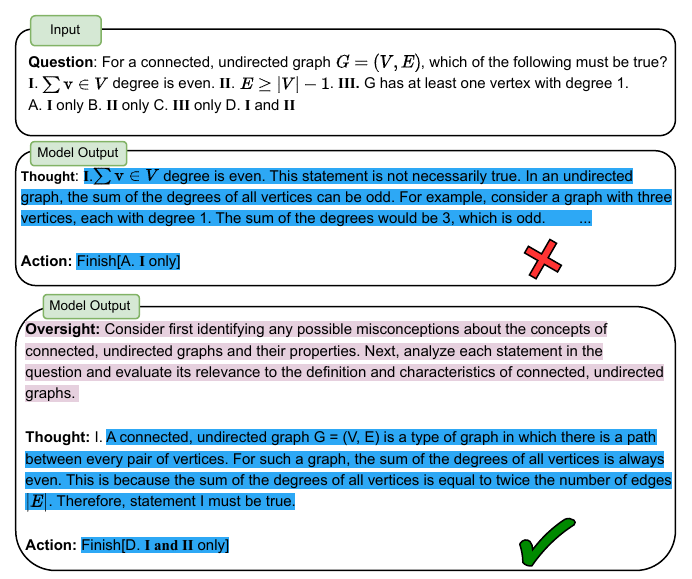}
    \caption{With question-oriented direction, the Reasoner answers questions with explicit clues.}
    \label{fig:inst_ex}
\end{figure}

\begin{figure}[t]
\centering
\begin{subfigure}[b]{0.48\textwidth}
   \includegraphics[width=1\linewidth,trim={13 0 0 0},clip]{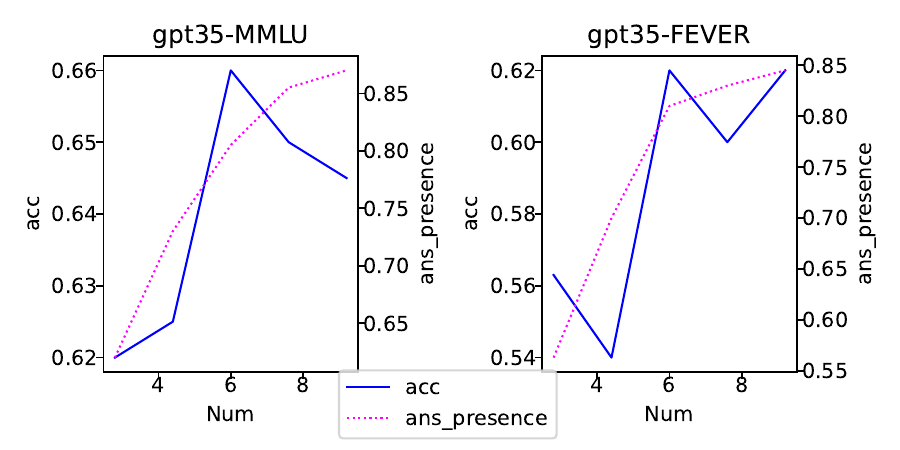}
   \caption{}
   \label{fig:acc_pres_other1} 
\end{subfigure}

\begin{subfigure}[b]{0.48\textwidth}
   \includegraphics[width=1\linewidth,trim={13 0 0 0},clip]{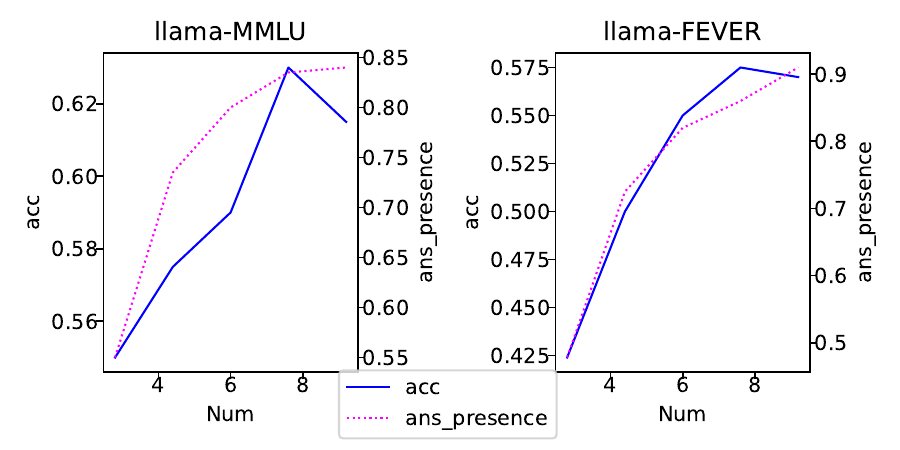}
   \caption{}
   \label{fig:acc_pres_other2}
\end{subfigure}
\caption{The task performance, Accuracy (acc) and the percentage of samples where the ground truth is included in the tree (ans-presence), with different size of search space (Num).}
\label{fig:searchSpace}
\end{figure}

\paragraph{The diversity of Search Space}
One of our motivations is to broaden the diversity of actions available for more effective exploration. Consequently, we compute the upper bound results for our generated tree, indicating the presence of the correct answer in the tree signifies a correctly answered sample. Results are shown in Figure~\ref{fig:searchSpace}. 

To analyse the effects of different LLMs quantitatively, we calculate the average pairwise semantic similarity between multiple directions for one question, then 1- similarity to obtain the diversity measurement shown below. The pretrained model, all-MiniLM-L6-v2~\footnote{\url{https://huggingface.co/sentence-transformers/all-MiniLM-L6-v2}} is used for sentence-pair similarity calculation.  The results is consistent with the intuition that sophisticated LLMs incline to less diverse instruction although such diverse directions are already capable to improve task performances.
\begin{table}[t]
    \centering
    \resizebox{0.48\textwidth}{!}{%
    \begin{tabular}{c|ccccc}
\toprule
&Stem&Social&Humanity&Other&Fever\\
 \midrule
GPT-3.5&0.665&0.673&0.674&0.648&0.632\\
LLama&0.689&0.691&0.688&0.669&0.661\\
Vicuna&0.694&0.690&0.692&0.679&0.673\\
\bottomrule
    \end{tabular}
    }
    \caption{Diversity among different directions generated by Mirror.}
    \label{tab:diversity}
\end{table}



\section{Ethics Statement}
We utilized two publicly available datasets: Massive Multitask Language Understanding (MMLU) and FEVER (Fact Extraction and Verification). MMLU is a multiple-choice question-answering dataset covering 57 subjects across STEM, social sciences, humanities, and more. Notably, some subjects, such as \textit{moral disputes} and \textit{moral scenarios}, contain statements that may raise ethical concerns. Here, LLMs could be misused or misinterpret the information. We strongly recommend thorough consideration of safety implications before applying such techniques in real-world scenarios. For the FEVER dataset, positive claims (facts) are extracted from Wikipedia, and negative claims are generated by contrasting these facts and subsequently verified without knowledge of their original source sentences. However, due to Wikipedia's editable nature, the extracted facts may not always be entirely accurate. Consequently, we advise against solely relying on our work as the truth source for any fact-checking task to avoid potential  confusion and bias.


\end{document}